\documentclass[journal]{IEEEtran}
\usepackage{latexsym,amsmath,amssymb,graphics,amscd, empheq, enumitem}
\usepackage[breaklinks,colorlinks,
  citecolor=blue,
  urlcolor=blue]{hyperref}
\usepackage{stackengine}

\usepackage{cite}
\ifCLASSINFOpdf
   \usepackage{graphicx}
\else
  \usepackage{graphicx}
\fi
\graphicspath{{figures/}}

\usepackage{amsthm}
\usepackage{soul }

\theoremstyle{plain}
\newtheorem{theorem}{Theorem}[section]
\newtheorem{proposition}[theorem]{Proposition}
\newtheorem{lemma}[theorem]{Lemma}
\newtheorem{corollary}[theorem]{Corollary}

\theoremstyle{definition}

\theoremstyle{remark}
\newtheorem{remark}[theorem]{Remark}

\newcommand{\vertiii}[1]{{\left\vert\kern-0.25ex\left\vert\kern-0.25ex\left\vert #1 
    \right\vert\kern-0.25ex\right\vert\kern-0.25ex\right\vert}}

\begin{document}

\title{Discrete-time signatures and randomness in reservoir computing}

\author{Christa Cuchiero, Lukas Gonon, Lyudmila Grigoryeva, Juan-Pablo Ortega, and Josef Teichmann
\thanks{C. Cuchiero is affiliated with the University of Vienna, Austria. L. Gonon is with the Ludwig-Maximilians-Universit\"at M\"unchen, Germany. L. Grigoryeva is with the Universit\"at Konstanz, Germany. J.-P. Ortega is with the  Universit\"at Sankt Gallen, Switzerland and the Centre National de la Recherche Scientifique (CNRS), France. J. Teichmann's affiliation is ETH Z\"urich, Switzerland. }
}

\markboth{Discrete-time signatures and randomness in reservoir computing}%
{Discrete-time signatures and randomness in reservoir computing}
%


\maketitle

\begin{abstract}
A new explanation of geometric nature of the reservoir computing phenomenon is presented. {Reservoir computing is understood in the literature as the possibility of approximating input/output systems with randomly chosen recurrent neural systems and a trained linear readout layer. Light is shed on this phenomenon by constructing what is called strongly universal reservoir systems as random projections of a family of state-space systems that generate Volterra series expansions.} This procedure yields a state-affine reservoir system with randomly generated coefficients in a dimension that is logarithmically reduced with respect to the original system. This reservoir system is able to approximate any element in the fading memory filters class just by training a different linear readout for each different filter. Explicit expressions for the probability distributions needed in the generation of the projected reservoir system are stated and bounds for the committed approximation error are provided.
\end{abstract}

\begin{IEEEkeywords}
Reservoir computing, recurrent neural network, state-affine system, SAS, signature state-affine system, SigSAS, echo state network, ESN, Johnson-Lindenstrauss Lemma, Volterra series, machine learning.
\end{IEEEkeywords}

%
\IEEEpeerreviewmaketitle

\section{Introduction}
%
%
%
%

\IEEEPARstart{M}{any}  dynamical problems in engineering, signal processing, forecasting, time series analysis, recurrent neural networks, or control theory can be described using {\it   input/output (IO) systems}. These mathematical objects establish a functional link that describes the relation between the time evolution of one or several explanatory variables (the input) and a second collection of dependent or explained variables   (the output). 

A generic question  in all those fields is to determine the IO system underlying an observed phenomenon. This is the so called {\it   system identification problem}. For this purpose, first principles coming from physics or chemistry can be invoked, when either these are known or the setup is simple enough to apply them. In complex situations, in which access to all the variables that determine the behavior of the systems is difficult or impossible, {or when a precise mathematical relation between input and output is not known,} it has proved more efficient to carry out the system identification using generic families of models with strong approximation abilities that are estimated using observed data. {This approach, that we refer to as {\it  empirical system identification}, has been developed using different techniques coming simultaneously from engineering, statistics, and computer science.}

In this paper, we focus on a particularly promising strategy for empirical system identification known as {\it   reservoir computing (RC)}. Reservoir computing capitalizes on the revolutionary idea that there are learning systems that attain universal approximation properties without the need to estimate all their parameters using, for instance,  supervised learning. More specifically, RC can be seen as a recurrent neural networks approach to model IO systems {\it using state-space representations in which} 
\begin{itemize}
\item 
{\it the state equation
is randomly generated, sometimes with sparsity features,} and
\item
{\it  only the  (usually very simple)  functional form of the
observation equation is tailored to the  specific problem
 using observed data.}
\end{itemize}
RC can be found in the literature under other denominations like {\it   Liquid State Machines}~\cite{Maass2000, maass1, Natschlager:117806, corticalMaass, MaassUniversality} and is represented by various learning paradigms, {with  {\it   Echo State Networks (ESNs)}  \cite{Matthews:thesis, Matthews1994, Jaeger04} being a} particularly important example.

RC has shown superior performance in many forecasting and classification engineering tasks (see \cite{lukosevicius, GHLO2014, Goudarzi2016, marzen:capacity} and references therein) and has shown unprecedented abilities in the learning of the attractors of complex nonlinear infinite dimensional dynamical systems \cite{Jaeger04, pathak:chaos, Pathak:PRL, Ott2018}. Additionally, RC implementations with dedicated hardware have been designed and built (see, for instance,~\cite{Appeltant2011, Rodan2011, SOASforRC, Larger2012, Paquot2012, photonicReservoir2013, swirl:paper, Vinckier2015, Laporte2018}) that exhibit information processing speeds that largely outperform standard Turing-type computers.

The most far-reaching and radical innovation in the RC approach is the use of untrained, randomly generated, and sometimes sparse state maps. {This circumvents well-known difficulties in the training of generic recurrent neural networks arising bifurcation }phenomena~\cite{Doya92}, which, despite recent progress in the regularization and training of deep RNN structures (see, for instance \cite{Graves2013, pascanu:rnn, zaremba}, and references therein), render classical gradient descent methods non-convergent. Randomization has already been successfully applied in a static setup using neural networks with randomized weights, in particular in seminal works on random feature models \cite{Rahimi2007} and Extreme Learning Machines \cite{Huang2006}. In the dynamic context, an important result in \cite{hart:ESNs} shows that randomly drawn ESNs can be trained by exclusively optimizing a linear readout using generic one-dimensional observations of a given invertible and differentiable dynamical system to produce dynamics that are topologically conjugate to that given system; in other words, randomly generated ESNs are capable of learning the attractors of invertible dynamical systems. More generally,  the approximation capabilities of randomly generated ESNs have been established in \cite{RC12} in the more general setup of IO systems. There, approximation bounds have been provided in terms of their architecture parameters.

In this paper, we provide an additional insight on the randomization question for another family of RC systems, namely, for the non-homogeneous state-affine systems (SAS).  These systems  have been introduced and proved to be universal
approximants in \cite{RC6, RC8}. We here show  that they also have this universality property when they are \emph{randomly} generated.
The approach pursued in this work is considerably different from the one in the above-cited references and is based on the following steps. First, we consider causal and time-invariant analytic filters with semi-infinite inputs. The Taylor series expansion of these objects coincide with what is known as their {\it    Volterra series representation}. Second, we show that the truncated Volterra series representation (whose associated truncation error can be quantified) admits a state-space representation with linear readouts in a (potentially) high-dimensional adequately constructed tensor space. We refer to this system as the {\it   signature state-affine system (SigSAS)}: on the one hand, it belongs to the SAS family  and, on the other hand, it shares two fundamental properties with the well-known signature process  from the theory of rough paths  \cite{chen1957integration, chen1958integration, Lyons:1998, hambly2010uniqueness}. First, the solutions of the SigSaS system fully characterize the input sequences and, second, any (sufficiently regular) IO system can be written as a linear map of the signature. These properties have been exploited in a continuous-time setup in \cite{JLPaper}.  Finally, we use the Johnson-Lindenstrauss Lemma \cite{JLlemma} to prove that a random projection of the SigSAS system yields a smaller dimensional SAS system with random matrix coefficients (that can be chosen to be sparse) that approximates the original
system. Moreover, this constructive procedure gives us full knowledge of the law that needs to be used to draw the entries of the low-dimensional SAS approximating system, without ever having to use the original large dimensional SigSAS, which amounts to a form of information compression with efficient reconstruction in this setup \cite{compressed:sensing:book}.  An important feature of the dimension reduced randomly drawn SAS system is that it serves as a universal approximator for any reasonably behaved IO system and that only the linear output layer that is applied to it depends on the individual system that needs to be learnt. We refer to this feature as the {\it   strong universality} property.

This approach to the approximation problem in recurrent neural networks using randomized systems provides a new explanation of geometric nature of the reservoir computing phenomenon. The results in the following pages show that {\it randomly generated  SAS reservoir systems approximate well any sufficiently regular IO system just by tuning a linear readout because they coincide with an error-controlled random projection of a higher dimensional Volterra series expansion of that system}.

\section{Truncated Volterra representations of analytic filters}

{We start by describing the setup that we shall be working on, together with the main approximation tool which we will be using  later on in the paper, namely, Volterra series expansions. Details on the concepts introduced in the following paragraphs can be found in, for instance, \cite{RC7, RC9}, and  references therein.}

{All along the paper, the symbol $\mathbb{Z}$ denotes the set of all integers and $\Bbb Z_-$ stands for the set of negative  integers with the zero element included}. Let $D_d \subset \mathbb{R}^d $ and $D_m \subset \mathbb{R}^m $. We  refer to the maps of the type $U: (D _d) ^{\Bbb Z} \longrightarrow (D_m) ^{\Bbb Z} $ between infinite sequences with values in $D _d $  and $D_m $, respectively,  as {\it   filters}, {\it   operators}, or discrete-time {\it   input/output systems}, and to those like $H: (D _d) ^{\Bbb Z} \longrightarrow D_m $ (or $H: (D _d) ^{\Bbb Z_-} \longrightarrow D_m $) as $\mathbb{R}^m $-valued  {\it   functionals}. These definitions will be sometimes extended to accommodate situations where the domains and the targets of the filters are not necessarily product spaces but just arbitrary subsets of $\left({\Bbb R}^d\right)^{\mathbb{Z}}  $ and $\left({\Bbb R}^m\right)^{\mathbb{Z}}  $ like, for instance, $\ell ^{\infty}(\mathbb{R}^d) $ and $\ell ^{\infty}(\mathbb{R}^m) $.

A filter $U: (D _d) ^{\Bbb Z} \longrightarrow (D_m) ^{\Bbb Z} $ is called {\it   causal} when for any two elements ${\bf z} , \mathbf{w} \in (D _d) ^{\Bbb Z}  $  that satisfy that ${\bf z} _\tau = \mathbf{w} _\tau$ for any $\tau \leq t  $, for a given  $t \in \Bbb Z $, we have that $U ({\bf z}) _t= U ({\bf w}) _t $. Let  $T_\tau:(D _d) ^{\Bbb Z} \longrightarrow(D _d) ^{\Bbb Z} $ be the {\it   time delay} operator defined by $T_\tau( {\bf z}) _t:= {\bf z}_{t- \tau}$. The filter $U$ is called {\it   time-invariant} (TI) when it commutes with the time delay operator, that is, $T_\tau \circ U=U  \circ T_\tau $,  for any $\tau\in \Bbb Z $ (in this expression, the two  operators $T_\tau $ have  to be understood as defined in the appropriate sequence spaces). There is a bijection between causal and time-invariant filters  and functionals. We denote by $U_H: (D _d) ^{\Bbb Z} \longrightarrow (D_m) ^{\Bbb Z} $ (respectively, $H_U: (D _d) ^{\Bbb Z_-} \longrightarrow D_m $) the filter (respectively, the functional) associated to the functional $H: (D _d) ^{\Bbb Z_-} \longrightarrow D_m $ (respectively, the filter $U: (D _d) ^{\Bbb Z} \longrightarrow (D_m) ^{\Bbb Z} $). Causal and time-invariant filters are fully determined by their restriction to semi-infinite sequences, that is, $U: (D _d) ^{\Bbb Z_-} \longrightarrow (D_m) ^{\Bbb Z_-} $, that will be denoted using the same symbol. 

In most cases, we work in the situation in which $D _d $ and $D _m $ are compact and the sequence spaces $(D _d) ^{\Bbb Z_-} $ and $(D_m) ^{\Bbb Z_-} $ are endowed with the product topology. It can be shown (see \cite{RC7}) that this topology is equivalent to the norm topology induced by any weighted norm defined by $\left\|{\bf z} \right\|_w :=\sup_{t \in \mathbb{Z}_{-}}\left\{ {\bf z} _t w_{-t}\right\}$, ${\bf z} \in (D _d) ^{\Bbb Z_-} $, where $w : \mathbb{N} \longrightarrow (0,1] $ is an arbitrary strictly decreasing sequence (we call it {\it   weighting sequence}) with zero limit and such that $w _0=1 $. Filters and functionals that are continuous with respect to this topology are said to have the {\it   fading memory property (FMP)}.

A particularly important class of IO systems are those generated by {\it   state-space sytems} in which the output ${\bf y} \in (D _m)^{\Bbb Z_-}$ is obtained out of the input ${\bf z} \in (D _d)^{\Bbb Z_-}$ as the solution of the equations
\begin{empheq}[left={\empheqlbrace}]{align}
\mathbf{x} _t &=F(\mathbf{x}_{t-1}, {\bf z} _t),\label{reservoir equation}\\
y _t &= h (\mathbf{x} _t), \label{readout}
\end{empheq}
where $F: D_N \times D_d \longrightarrow D_N $ is the so-called {\it   state map}, for some $D_N\subset \mathbb{R}^N  $, $N \in \mathbb{N} $,  and $h: D _N \longrightarrow D_m $  is the {\it   readout} or {\it   observation} map. When for any input ${\bf z} \in (D _d) ^{\Bbb Z_-} $ there is only one output ${\bf y}\in (D _m)^{\Bbb Z_-} $ that satisfies \eqref{reservoir equation}-\eqref{readout}, we say that this state-space system has the {\it   echo state property (ESP)}, in which case it determines a unique filter $U^F _h: (D _d) ^{\Bbb Z_-} \longrightarrow (D_m) ^{\Bbb Z_-} $. When the ESP holds at the level of the state equation \eqref{reservoir equation}, then it determines another filter $U^F: (D _d) ^{\Bbb Z_-} \longrightarrow (D_N) ^{\Bbb Z_-} $ and then $U^F_h= h(U^F) $. The filters $U^F_h $ and $U^F$, when they exist,  are automatically causal and TI (see \cite{RC7}). The continuity and the differentiability properties of the state and observation maps $F $ and $h$ imply continuity and differentiability for $U^F_h $ and $U^F$ under very general hypotheses; see \cite{RC9} for an in-depth study of this question.

We denote by $\|{\cdot}\|$ the Euclidean norm if not stated otherwise and use the symbol $\vertiii{\cdot}$ for the operator norm with respect to the $2$-norms in the target and the domain spaces. Additionally, for  any ${\bf z} \in (\mathbb{R}^d)^{\mathbb{Z}_{-}}$ we define $p$-norms as $\left\|{\bf z}\right\|_ p := \left( \sum_{t\in \mathbb{Z}_-} \|{\bf z}_t \|^p \right)^{1/p}  $,  for $1 \le p <\infty $, and  $\left\|{\bf z}\right\|_ \infty:=\sup_{t \in \mathbb{Z}_{-}}  \left\{\left\|{\bf z} _t\right\|\right\}$ for $p=\infty$. Given $M>0 $, we denote by $K _M:= \{{\bf z} \in (\mathbb{R}^d)^{\mathbb{Z}_{-}}\mid \left\|{\bf z} _t\right\|\leq M \enspace {\rm for} \enspace {\rm all} \enspace t\in \Bbb Z _-\}$. It is easy to see that $K_M = \overline{B _M}
\subset \ell_{-}^{\infty}(\mathbb{R}^d)$, with 
$
B _M:=B_{\left\|\cdot \right\|_\infty}({\bf 0}, M)$  and $\ell_{-}^{\infty}(\mathbb{R}^d):= \{{\bf z} \in (\mathbb{R}^d)^{\mathbb{Z}_{-}}\mid \left\|{\bf z}\right\|_ \infty< \infty \}$. {Additionally, let $\widetilde{B}_M:=B_M\cap \ell_{-}^{1}(\mathbb{R}^d)$ with $\ell_{-}^{1}(\mathbb{R}^d):= \{{\bf z} \in (\mathbb{R}^d)^{\mathbb{Z}_{-}}\mid \left\|{\bf z}\right\|_ 1< \infty \}$. We will use the same symbol $\widetilde{B}_M$ whenever $d=1$ in the sequel. The following statement  is the main approximation result that will be used in the paper. }

\begin{theorem}
\label{volterra approximation}
Let $M, L>0 $ and let $U:K _M \subset \ell_{-}^{\infty}(\mathbb{R}^d) \longrightarrow K _L \subset \ell_{-}^{\infty}(\mathbb{R}^m) $ be a causal and time-invariant fading memory filter whose restriction $U|_{B _M}$ is analytic as a map between open sets in the Banach spaces $\ell_{-}^{\infty}(\mathbb{R}^d) $ and  $\ell_{-}^{\infty}(\mathbb{R}^m)$ and satisfies $U({\bf 0})= {\bf 0} $. Then, {for any ${\bf z}\in\widetilde{B}_M$} there exists a Volterra series representation of $U$ given by
\begin{equation}
\label{volterra representation}
\footnotesize
U ({\bf z})_t=\sum_{j=1}^{\infty}\sum_{m _1=- \infty}^0 \cdots\sum_{m _j=- \infty}^0 g _j(m _1, \ldots, m _j)({\bf z}_{ m _1+ t}\otimes \cdots \otimes{\bf z}_{ m _j+ t}),
\end{equation}
with $ t \in \mathbb{Z}_{-}$ and where the map $g _j: (\Bbb Z_-) ^j \longrightarrow L({\Bbb R}^d \otimes \cdots \otimes{\Bbb R}^d, {\Bbb R}^m) $ is given by
\begin{equation}
\label{definition g maps}
g _j(m _1, \ldots, m _j)({\rm \mathbf{e}}_{i _1}\otimes \cdots \otimes{\rm \mathbf{e}}_{i _j})= \frac{1}{j!}D ^jH _U({\bf 0})({\rm \mathbf{e}}^{i _1}_{m _1}, \ldots ,{\rm \mathbf{e}}^{i _j}_{m _j}),
\end{equation}
where $\left\{\mathbf{e}_1, \ldots, \mathbf{e}_d\right\}$ is the canonical basis of $\mathbb{R}^d  $ and the sequences ${\rm \mathbf{e}}^{i _l}_{m _k}\in \ell_{-}^{\infty}(\mathbb{R}^d)$ are defined by:
\begin{equation*}
\left({\rm \mathbf{e}}^{i _l}_{m _k}\right)_t:=
\left\{
\begin{array}{cr}
\mathbf{e}_{i _l}\in {\Bbb R}^d, &{\rm if}\ t=m _k,\\
{\bf 0},& {\rm otherwise.}
\end{array}
\right.
\end{equation*}
Moreover, there exists a monotonically decreasing sequence $w ^U $ with zero limit such that, for any $p,l \in \mathbb{N} $,
\footnotesize
\begin{multline}
\label{volterra representation truncated}
\!\!\!\!\!\!\!\!\!\!\left\|U ({\bf z})_t- \sum_{j=1}^{p}\sum_{m _1=- l}^0 \cdots\sum_{m _j=- l}^0 g _j(m _1, \ldots, m _j)({\bf z}_{ m _1+ t}\otimes \cdots \otimes{\bf z}_{ m _j+ t})\right\|\\
\leq w ^U _l+L \left(1- \frac{\left\|{\bf z}\right\|_{\infty}}{M}\right)^{-1} \left(\frac{\left\|{\bf z}\right\|_{\infty}}{M}\right)^{p+1}.
\end{multline}
\normalsize
\end{theorem}

\subsection{The signature state-affine system (SigSAS)}
\label{The signature state-affine system (SigSAS)}

We now show that the filter obtained out of the truncated Volterra series expansion in the expression \eqref{volterra representation truncated} can be written down as the unique solution of a non-homogeneous state-affine system (SAS) with linear readouts that, as we shall show in Section \ref{The SigSAS approximation theorem}, has particularly strong universal approximation properties. We first briefly recall how the SAS family is constructed. 

Let $\boldsymbol{\alpha} = (\alpha_1, \ldots, \alpha_d)^{\top} \in \mathbb{N}^d$ and ${\bf z} = (z_1, \ldots, z_d)^{\top} \in \mathbb{R}^d$. We define the monomials ${\bf z}^{\boldsymbol{\alpha}} := z_1^{\alpha_1} \cdots  z_d^{\alpha_d}$. Let $N_1,N_2 \in \mathbb{N}  $  and let $\mathbb{M}_{N_1,N_2 } [{\bf z }]$ be the space of polynomials in ${\bf z} \in \mathbb{R}^d$ with matrix coefficients in $\mathbb{M}_{N_1,N_2 }$, 
that is, the set of elements $p$ of the form
\begin{equation*}
    p({\bf z}) = \sum_{\boldsymbol{\alpha} \in V_p }{\bf z}^{\boldsymbol{\alpha}} A_{\boldsymbol{\alpha}},
\end{equation*}
with $V_p \subset \mathbb{N}^d$ a finite subset and $A_{\boldsymbol{\alpha}}\in\mathbb{M}_{N_1,N_2 }$ the matrix coefficients.  A {\it  state-affine system (SAS)} is given by
\begin{equation}
\label{state-affine definition}
    \begin{cases}
        {\bf x}_t = p({\bf z}_t) {\bf x}_{t-1} + q({\bf z}_t),\\
        {\bf y}_t = W {\bf x}_t,
    \end{cases}
\end{equation}
 { $p \in \mathbb{M}_{N, N}[{\bf z}]$, $q \in \mathbb{M}_{N, 1}[{\bf z}]$ are polynomials with matrix and vector coefficients, respectively}, and $W \in \mathbb{M}_{m, N}$. If we consider inputs in the set $K _M $ and the $p$ and $q$ in the state-space system \eqref{state-affine definition} such that
 \begin{eqnarray*}
     M_p &:=& \sup_{{\bf z} \in \overline{B_{\left\|\cdot \right\|}({\bf 0}, M)}} \left \{ \vertiii{p({\bf z})} \right \} < 1,\\
M_q &:=& \sup_{{\bf z} \in \overline{B_{\left\|\cdot \right\|}({\bf 0}, M)}} \left \{ \vertiii{q({\bf z})} \right \} < \infty,
 \end{eqnarray*}
where $\overline{B_{\left\|\cdot \right\|}({\bf 0}, M)}$ denotes the closed ball in $\mathbb{R}^d$ of radius $M$ and center  $\mathbf{0}$ with respect to the Euclidean norm,
then a unique state-filter $U^{p,q}:K _M \longrightarrow K_L$ can be associated to it, with $L:=M _q/(1-M _p)$. It has been shown in \cite{RC6, RC8} that SAS systems are universal approximants in the fading memory and in the $L ^p $-integrable categories in the sense that given a filter in any of those two categories, {there exists a SAS system of type \eqref{state-affine definition} that uniformly {or in the $L ^p$-sense} approximates it}.

The signature state-affine system that we construct in this section exhibits what we call the {\it   strong universality property}. This means that the state equation for this state-space representation is {\it the same} for any fading memory filter that is being approximated, and it is only the linear readout that changes. In other words, we provide a result that yields the approximation (as accurate as desired) of any fading memory IO system, as the linear readout of the solution of a fixed non-homogeneous SAS system that {\it does not depend on the filter being approximated}. 

Since the important  property that we just described is reminiscent of an analogous feature of the signature process in the context of the representation of the solutions of controlled stochastic differential equations \cite{JLPaper}, we shall refer to this state system as the {\it   signature SAS (SigSAS)} system.

Before we proceed, we need to introduce some notation. First, for any $l,d \in \mathbb{N} $, we denote by $T ^l( \mathbb{R} ^d)$ the space of tensors of order $l$ on $\mathbb{R} ^d $, that is,
\begin{equation*}
T ^l( \mathbb{R} ^d):= \left\{\sum_{i _1, \ldots, i _l=1}^d a_{i _1, \ldots, i _l}\mathbf{e}_{i _1}\otimes \cdots \otimes\mathbf{e}_{i _l}\mid a_{i _1, \ldots, i _l} \in \mathbb{R}\right\}.
\end{equation*}
The tensor space $T ^l( \mathbb{R} ^d) $ will be {understood as} a normed space with a crossnorm  \cite{Lancaster1972} that we shall leave unspecified for the time being. We shall be using an {\it   order lowering map} $\pi_{l}: T ^{l+1}( \mathbb{R} ^d)  \longrightarrow T ^{l}( \mathbb{R} ^d) $ that, for any vector $\mathbf{v}:=\sum_{i _1, \ldots, i _{l+1}=1}^d a_{i _1, \ldots, i _{l+1}}\mathbf{e}_{i _1}\otimes \cdots \otimes\mathbf{e}_{i _{l+1}} \in  T ^{l+1}( \mathbb{R} ^d)$,  is defined as, 
\begin{equation*}
\pi_l( \mathbf{v}):=\sum_{i _2, \ldots, i _{l+1}=1}^d a_{1,i _2, \ldots, i _{l+1}}\mathbf{e}_{i _2}\otimes \cdots \otimes\mathbf{e}_{i _{l+1}} \in  T ^{l}( \mathbb{R} ^d).
\end{equation*}
The order lowering map is linear and its operator norm satisfies that $\vertiii{\pi_{l}}=1 $.

We shall restrict the presentation to one-dimensional inputs, that is, we consider input sequences ${\bf z} \in K _M\subset \ell_{-}^{\infty}(\mathbb{R}) $. Now, for fixed $l,p \in \mathbb{N} $, we define for any ${\bf z} \in K _M $ and $t \in \mathbb{Z}_{-}$,
\begin{equation}
\label{ztilde and zhat}
\widetilde{{\bf z}} _t:=\sum _{i=1}^{p+1} z _t^{i-1}\mathbf{e} _i \in \mathbb{R} ^{p+1} \quad \mbox{and} \quad
\widehat{{\bf z}} _t:=\widetilde{{\bf z}} _{t-l} \otimes \cdots \otimes \widetilde{{\bf z}} _t.
\end{equation}
Note that $\widetilde{{\bf z}} _t $ is the  {\it   Vandermonde vector} \cite{hankel:tensors} associated to $z _t $ and that $\widehat{{\bf z}} _t $ is a tensor in $T^{l+1}\left(\mathbb{R}^{p+1}\right) $ whose components in the canonical basis are all the monomials on the variables $z _t, \ldots, z _{t-l}$ that contain powers up to order $p$ in each of those variables. Indeed,
\begin{equation*}
\widehat{{\bf z}} _t=\sum_{i _1, \ldots, i _{l+1}=1}^{p+1} z_{t-l}^{i_1-1} \cdots z_{t}^{i_{l+1}-1}\mathbf{e}_{i _1}\otimes \cdots \otimes\mathbf{e}_{i _{l+1}} \in T^{l+1}\left(\mathbb{R}^{p+1}\right).
\end{equation*}
Finally, given $I _0 \subset \left\{1, \ldots, p+1\right\} $ a subset of cardinality higher than $1$ that contains the element 1, we define:
\begin{equation}
\label{definition z0}
\widehat{{\bf z}} _t^0=\sum_{i \in I _0} z_{t}^{i-1}\underbrace{\mathbf{e}_{1}\otimes \cdots \otimes\mathbf{e}_{1}}_{\text{$l$-times}}\otimes\mathbf{e}_{i} \in T^{l+1}\left(\mathbb{R}^{p+1}\right).
\end{equation}
The next proposition introduces the SigSAS state system for fixed $l, p \in \mathbb{N}  $, whose {solution is} used later on in Theorem \ref{SigSAS approximation} to represent the truncated Volterra series expansions in Theorem \ref{volterra approximation} of polynomial degree $p$  and lag $-l $ (see expression \eqref{volterra representation truncated}).

\begin{proposition}[The SigSAS system]
\label{proposition The SigSAS system}
Let $M>0 $  and let $l,p \in \mathbb{N} $. Let $0< \lambda < \min \left\{1, \frac{1-M}{1-M^{p+1}}\right\} $. Consider the state system with uniformly bounded scalar inputs in $K _M = [-M,M]^{\mathbb{Z}_{-}} $ and states in $T^{l+1}(\mathbb{R}^{p+1})$ given by the recursion
\begin{equation}
\label{sigsas recursion}
\mathbf{x} _t= \lambda \pi_l (\mathbf{x}_{t-1})\otimes \widetilde{{\bf z}} _t+ \widehat{{\bf z}} _t^0.
\end{equation}
This state equation is induced by the state map $F^{{\rm SigSAS}}_{\lambda,l , p}:T^{l+1}(\mathbb{R}^{p+1}) \times \mathbb{R} \longrightarrow T^{l+1}(\mathbb{R}^{p+1}) $ defined by
\begin{equation}
\label{sigsas state function}
F^{{\rm SigSAS}}_{\lambda,l , p}(\mathbf{x}, z):=\lambda \pi_l (\mathbf{x})\otimes \widetilde{{\bf z}} + \widehat{{\bf z}} ^0,
\end{equation}
which is a contraction in the state variable with contraction constant
\begin{equation}
\label{contraction constant}
\lambda \widetilde{M} <1, \quad \mbox{where} \quad \widetilde{M}:=\frac{1-M^{p+1}}{1-M}=\sum_{j=0}^p M ^j,
\end{equation}
and hence restricts to a map $F^{{\rm SigSAS}}_{\lambda,l , p}: \overline{B _{\left\|\cdot \right\|}({\bf 0},L)} \times [-M, M] \longrightarrow \overline{B _{\left\|\cdot \right\|}({\bf 0},L)}  $, with 
\begin{equation}
\label{definition of L}
L:= \widetilde{M}/ (1- \lambda \widetilde{M}).
\end{equation}
This state system has  the echo state and the fading memory properties and its  continuous, time-invariant, and causal associated filter $U^{{\rm SigSAS}}_{\lambda,l , p}: K _M \longrightarrow K _L \subset T^{l+1}(\mathbb{R}^{p+1})$ is given by:
\footnotesize
\begin{multline}
\label{solution sigsas}
U^{{\rm SigSAS}}_{\lambda,l , p}({\bf z})_t= \frac{\lambda^{l+1}}{1- \lambda}\widehat{{\bf z}} _t\\
+ \lambda ^l
\underbrace{\pi _l( \pi_l( \cdots (\pi_l(}_\text{$l$-times}\widehat{{\bf z}}^0_{t-l})\otimes\widetilde{{\bf z}}_{t-(l-1)}) \otimes \cdots )\otimes \widetilde{{\bf z}}_{t-1}) \otimes\widetilde{{\bf z}}_{t}+\\
 \cdots+ \lambda \pi_l(\widehat{{\bf z}}^0_{t-1})\otimes \widetilde{ {\bf z}} _t+ \widehat{{\bf z}}^0_{t}.
\end{multline}
\normalsize
\end{proposition}

\begin{remark}
\normalfont
The state equation \eqref{sigsas recursion} is indeed a SAS with states defined in $T^{l+1}(\mathbb{R}^{p+1})  $ as it has the same form as the first equality in \eqref{state-affine definition}. Indeed, this equation can be written as $ {\bf x}_t = p({  z}_t) {\bf x}_{t-1} + q({ z}_t)$ with $p({  z}_t) $ and $q({  z}_t)$ the polynomials in $z _t $ with coefficients in $L(T^{l+1}(\mathbb{R}^{p+1}), T^{l+1}(\mathbb{R}^{p+1}))$ and $T^{l+1}(\mathbb{R}^{p+1}) $, respectively, given by:
\begin{eqnarray*}
p(z _t) \mathbf{x}_{t-1}&:= & \lambda \pi_l (\mathbf{x}_{t-1})\otimes \widetilde{{\bf z}} _t=\sum_{i=1}^{p+1}z _t^{i-1}\left(\lambda \pi_l(\mathbf{x}_{t-1})\otimes \mathbf{e}_i\right),\\
q(z _t)&:= &\widehat{{\bf z}} _t^0=\sum_{i \in I _0} z_{t}^{i-1}\mathbf{e}_{1}\otimes \cdots \otimes\mathbf{e}_{1}\otimes\mathbf{e}_{i}.
\end{eqnarray*}
\end{remark}

\subsection{The SigSAS approximation theorem}
\label{The SigSAS approximation theorem}

As we already pointed out, $\widehat{{\bf z}} _t $ is a vector in $T^{l+1}\left(\mathbb{R}^{p+1}\right) $ whose components in the canonical basis are all the monomials on the variables $z _t, \ldots, z _{t-l}$ that contain powers up to order $p$ in each of those variables. Moreover, it is easy to see that all the other summands in the expression \eqref{solution sigsas} of the filter $U^{{\rm SigSAS}}_{\lambda,l , p}$ are proportional (with a positive constant) to monomials already contained in $\widehat{{\bf z}} _t $. This implies the existence of a {linear map $A_{\lambda,l , p} \in L(T^{l+1}(\mathbb{R}^{p+1}), T^{l+1}(\mathbb{R}^{p+1}))$ with an invertible matrix representation with non-negative entries such that 
\begin{equation}
\label{transformation for monomials}
U^{{\rm SigSAS}}_{\lambda,l , p}({\bf z})_t =A_{\lambda,l , p}  \widehat{ {\bf z}} _t.
\end{equation}
In the sequel we will denote the matrix representation of $A_{\lambda,l , p}$ using  the same symbol $A_{\lambda,l , p} \in \mathbb{M}_{N,N}$, $N := (p+1)^{l+1}$.}
This observation, together with Theorem \ref{volterra approximation}, can be used to prove the following result.

\begin{theorem}
\label{SigSAS approximation}
Let $M, L>0 $ and let $U:K _M \subset  \ell_{-}^{\infty}(\mathbb{R})\longrightarrow K_L \subset \ell_{-}^{\infty}(\mathbb{R}^m) $ be a causal and time-invariant fading memory filter whose restriction $U|_{B _M}$ is analytic as a map between open sets in the Banach spaces $\ell_{-}^{\infty}(\mathbb{R}) $ and $\ell_{-}^{\infty}(\mathbb{R}^m) $ and satisfies $U({\bf 0})= {\bf 0} $. Then, there exists a monotonically decreasing sequence $w ^U $ with zero limit such that, for any $p,l \in \mathbb{N} $, and any $0< \lambda < \min \left\{1, \frac{1-M}{1-M^{p+1}}\right\} $, there exists a linear map $W \in L(T^{l+1}(\mathbb{R}^{p+1}), \mathbb{R} ^m)$ such that,  for any ${\bf z}\in\widetilde{B}_M$:
\footnotesize
\begin{equation}
\label{volterra representation truncated sigsas}
\left\|U ({\bf z})_t- WU^{{\rm SigSAS}}_{\lambda,l , p}({\bf z})_t \right\|
\leq w ^U _l+L \left(1- \frac{\left\|{\bf z}\right\|_{\infty}}{M}\right)^{-1} \left(\frac{\left\|{\bf z}\right\|_{\infty}}{M}\right)^{p+1}.
\end{equation}

\normalsize
\end{theorem}

\begin{remark}
\label{strong but}
\normalfont
Theorem \ref{SigSAS approximation} establishes the strong universality of the SigSAS system in the sense that the state equation of this system is { the same} for any fading memory filter $U$ that is being approximated, and it is only the linear readout that changes. Nevertheless, we emphasize that the quality of the approximation {\it is not} filter independent, as the decreasing sequence $w ^U $ in the bound \eqref{volterra representation truncated sigsas} depends on how fast the filter $U$ ``forgets" past inputs.
\end{remark}

\begin{remark}
\normalfont
The analyticity hypothesis in the statement of Theorem \ref{SigSAS approximation} can be dropped by using the fact that finite order and finite memory Volterra series are universal approximators in the fading memory category (see \cite{Boyd1985} and \cite[Theorem 31]{RC9}). In that situation, the bound for the truncation error in \eqref{volterra representation truncated sigsas} does not necessarily apply anymore, in particular its second summand, which is intrinsically linked to analyticity. A generalized bound can be formulated in that case using arguments along the lines of those found in \cite{RC12}.
\end{remark}

\section{Johnson-Lindenstrauss  reduction of the SigSAS representation}

The price to pay for the strong universality property exhibited by the signature state-affine system that we constructed in the previous section, is the potentially large dimension of the tensor space in which this state-space representation is defined. In this section we concentrate on this problem by proposing a dimension reduction strategy {which consists in using  the random projections in the Johnson-Lindenstrauss Lemma \cite{JLlemma} in order to construct} a smaller dimensional SAS system with random matrix coefficients (that can be chosen to be sparse). The results  contained in the next  subsections quantify the increase in approximation error committed when applying this dimensionality reduction strategy.

We start by introducing the Johnson-Lindenstrauss (JL) Lemma \cite{JLlemma}  and some properties that are needed later on in the presentation. Following this, we spell out how to use it in the dimension reduction of state-space systems in general and of the SigSAS representation in particular.

\subsection{The JL Lemma and approximate projections}

Given a $N$-dimensional Hilbert space $ \left(V, \langle\cdot , \cdot \rangle\right)$ and $Q$ a $n$-point subset of $V$, the {\it Johnson-Lindenstrauss (JL) Lemma} \cite{JLlemma} guarantees, for any $0< \epsilon<1 $, the existence of a linear map $f:V \longrightarrow \mathbb{R}^k $, with $k\in \mathbb{N}$ satisfying
\begin{equation}
\label{JL dimension condition}
k\geq \frac{24 \log n}{3 \epsilon ^2-2 \epsilon^3},
\end{equation}
that respects $\epsilon$-approximately the distances between the points in the set $Q$. More specifically,
\begin{equation}
\label{JL distances condition}
(1- \epsilon) \left\|\mathbf{v} _1- \mathbf{v} _2\right\| ^2\leq  \left\|f(\mathbf{v} _1)- f(\mathbf{v} _2)\right\| ^2\leq (1+ \epsilon) \left\|\mathbf{v} _1- \mathbf{v} _2\right\| ^2,
\end{equation}
for any $\mathbf{v} _1, \mathbf{v}_2 \in Q $. The norm $ \left\|\cdot \right\| $ in $\mathbb{R}^k  $ comes from an inner product that makes it into a Hilbert space or, in other words, it satisfies the parallelogram identity. This remarkable result is even more so in connection with further developments that guarantee that the linear map $f$ can be randomly chosen \cite{frankl1988johnson, indyk1998approximate, dasgupta2003elementary} and, moreover, within a family of sparse transformations \cite{achlioptas2003database, dasgupta2010sparse} (see also \cite{matouvsek2008variants}).

In the developments in this paper, we use the original version of this result in which the JL map $f$ is realized by a matrix $A \in \mathbb{M}_{k,N} $ whose entries are such that 
\begin{equation}
\label{choice for JL 1/k}
A _{ij}\sim {\rm N}(0, 1/k). 
\end{equation}
It can be shown that with this choice, the probability of the relation \eqref{JL distances condition} to hold for any pair of points in $Q$ is bounded below by {${1}/{n} $}.

\begin{lemma}
\label{definition of Q norm}
Let $(V, \left\|\cdot \right\|)$ be a normed space and  let $Q$ be a (finite or infinite {countable}) subset of $V$. Define $\left\|\cdot \right\| _Q: {\rm span}\left\{Q\right\}\longrightarrow \mathbb{R}_+ $ by
\begin{equation*}
\left\|\mathbf{v}\right\|_Q:= {\rm inf} \left\{\sum_{j=1}^{{\rm Card} \, Q }| \lambda _j | \, \Big | \sum_{j=1}^{{\rm Card} \, Q } \lambda _j \mathbf{v} _j= \mathbf{v}, \mathbf{v} _j \in Q\right\}.
\end{equation*}
\begin{description}[leftmargin=1.5em]
\item [(i)]  $\left\|\cdot \right\| _Q $  defines a seminorm in ${\rm span}\left\{Q\right\} $. If 
\begin{equation}
\label{definition of M}
M_Q:=\sup \left\{\left\|\mathbf{v} _i\right\|\mid \mathbf{v} _i\in Q\right\} 
\end{equation}
is finite, then $\left\|\cdot \right\| _Q $ is  a norm.
\item [(ii)] $\left\|\mathbf{v}\right\|\leq \left\|\mathbf{v}\right\|_Q M_Q $, for any $\mathbf{v}  \in {\rm span}\left\{Q\right\} $.
\item [(iii)] Let $Q _1, Q _2 $ be subsets of $V$ such that  $Q _1\subset Q _2 $. Then $\left\|\mathbf{v}\right\|_{Q _2}\leq \left\|\mathbf{v}\right\|_{Q _1}$ for any $\mathbf{v} \in {\rm span}\left\{Q_1\right\}$.
\end{description}
\end{lemma}

\begin{remark}
\normalfont
If the hypothesis $M_Q< \infty $ is dropped in part {\bf (i)} of Lemma \ref{definition of Q norm}, then $\left\|\cdot \right\|_Q $ is in general not a norm as the following example shows. Take $V= \mathbb{R}  $ and $\mathbf{v} _i=i $, $i \in \mathbb{N}  $. It is easy to see that, in this setup,
\begin{equation*}
\left\|1\right\|_Q=\inf \left\{\frac{1}{i}\mid i \in \mathbb{N}\right\}=0.
\end{equation*}
\end{remark}

\begin{proposition}
\label{quasi projection proposition}
Let $Q$ be a set of points in the Hilbert space $(V, \langle\cdot , \cdot \rangle) $ with $M_Q:=\sup \left\{\left\|\mathbf{v} _i\right\|\mid \mathbf{v} _i\in Q\right\} < \infty$ such that   $-Q:= \left\{- \mathbf{v}\mid \mathbf{v} \in Q\right\} =Q$. Let $\epsilon>0  $, let $f:V \longrightarrow \mathbb{R}^k $ be a linear map that satisfies the Johnson-Lindenstrauss property \eqref{JL distances condition} with respect to $\epsilon $, and let $f ^\ast : \mathbb{R}^k\longrightarrow V $ the adjoint map with respect to a fixed inner product $\langle\cdot , \cdot \rangle $ in $\mathbb{R}^k $. Then,
\begin{equation}
\label{epsilon condition ffstar}
\left| \langle\mathbf{w} _1, \left(\mathbb{I} _V- f ^\ast \circ f\right) (\mathbf{w} _2)\rangle\right| \leq \epsilon M_Q ^2 \left\|\mathbf{w} _1\right\|_Q \left\|\mathbf{w} _2\right\|_Q,
\end{equation}
for any $\mathbf{w}_1, \mathbf{w} _2 \in {\rm span}\, \{Q \}$. 
\end{proposition}

\begin{corollary}
In the hypotheses of the previous proposition, let 
\begin{equation}
\label{definition of C_Q}
C _Q:=\inf_{c \in \mathbb{R}_+} \left\{\left\|\mathbf{v}\right\|_Q\leq c \left\|\mathbf{v}\right\| \mbox{, for all $\mathbf{v} \in {\rm span}\{ Q\} $}\right\}.
\end{equation}
Then, for any $\mathbf{v} \in {\rm span}\{ Q\} $ {such that $(f ^\ast \circ f)(\mathbf{v}) \in {\rm span}\{ Q\} $}, we have
\begin{equation}
\label{inequ with norm cq}
\left\|\left(\mathbb{I} _V- f ^\ast \circ f\right) (\mathbf{v})\right\|\leq \epsilon M_Q ^2C _Q^2 \left\|\mathbf{v}\right\|.
\end{equation}
\end{corollary}

This corollary is just a consequence of the inequality \eqref{epsilon condition ffstar} that guarantees that
\begin{multline}
\label{variation for later}
\left\|\left(\mathbb{I} _V- f ^\ast \circ f\right) (\mathbf{v})\right\|^2\leq \epsilon M_Q ^2\left\|\left(\mathbb{I} _V- f ^\ast \circ f\right) (\mathbf{v})\right\| _Q \left\|\mathbf{v}\right\|_Q\\
\leq 
\epsilon M_Q ^2 C _Q ^2\left\|\left(\mathbb{I} _V- f ^\ast \circ f\right) (\mathbf{v})\right\|  \left\|\mathbf{v}\right\|,
\end{multline}
which yields \eqref{inequ with norm cq}.

\subsection{Johnson-Lindenstrauss projection of state-space dynamics}

The next result shows how, when the dimension $k$ of the target of the JL map $f$ determined by \eqref{JL dimension condition} is chosen so that this map is generically surjective, then any contractive state-space system with states in the domain of $f$ can be projected onto another one with states in its smaller dimensional image. This result also shows that if the original system has the ESP and the FMP, then so does the projected one. Additionally, it gives bounds that quantify the dynamical differences between the two systems. 

\begin{theorem}
\label{theorem Johnson-Lindenstrauss projection of state-space dynamics}
Let $F _\rho: \mathbb{R}^N \times D _d \longrightarrow  \mathbb{R}^N $ be a one-parameter family of continuous state maps, where $D_d \subset \mathbb{R} ^d $ is a compact subset,  $0< \rho< 1 $, and $F _\rho  $ is a  $\rho $-contraction on the first component. Let $Q$ be a $n$-point spanning subset of $
\mathbb{R}^N  $  satisfying $-Q=Q $. Let $f : \mathbb{R}^N \longrightarrow \mathbb{R}^k  $ be a JL map that satisfies \eqref{JL distances condition} with $0< \epsilon<1 $ where the dimension $k$ has been chosen so that $f$ is generically surjective. Then:
\begin{description}[leftmargin=1.5em]
\item [(i)] Let $F^f _\rho: \mathbb{R}^k \times D _d \longrightarrow \mathbb{R}^k $ be the state map defined by:
\begin{equation*}
F^f _\rho(\mathbf{x}, {\bf z}):= f \left(F _\rho(f ^\ast (\mathbf{x}), {\bf z})\right), 
\end{equation*}
for any $\mathbf{x} \in \mathbb{R}^k $ and ${\bf z} \in D _d $.
If the parameter $\rho $  is chosen so that 
\begin{equation}
\label{esp condition in f}
\rho< 1/\vertiii{f} ^2,  
\end{equation}
then $F^f _\rho$ is a contraction on the first entry. The symbol $\vertiii{\cdot }$ in \eqref{esp condition in f} denotes the operator norm with respect to the 2-norms in $\mathbb{R}^N  $ and $\mathbb{R} ^k $.
\item [(ii)] Let $V_k:= f ^\ast (\mathbb{R} ^k)\subset \mathbb{R} ^N$ and let $\mathcal{F}^f _\rho:V_k \times D _d\longrightarrow V_k$ be the state map with states on the vector space $V _k  $, defined by:
\begin{equation}
\label{definition frho}
\mathcal{F}^f _\rho(\mathbf{x}, {\bf z}):= f^\ast  \left(F^f _\rho((f ^\ast) ^{-1} (\mathbf{x}), {\bf z})\right)= f ^\ast \circ  f  (F_\rho(\mathbf{x}, {\bf z})), 
\end{equation}
for any $\mathbf{x} \in V_k $ and ${\bf z} \in D _d $.
If the contraction parameter satisfies \eqref{esp condition in f} then $\mathcal{F}^f _\rho $ is also a contraction on the first entry. Moreover, the restricted linear map $f ^\ast : \mathbb{R} ^k \longrightarrow V_k $ is a state-map equivariant linear isomorphism between $F ^f _\rho $  and $\mathcal{F} ^f _\rho $. 
\item [(iii)] Suppose, additionally, that there exist two constants  $C, C _f>0 $ such that the state spaces of the state maps $F _\rho $ and $F _\rho^f  $ can be restricted as $F _\rho: \overline{B_{\left\| \cdot \right\|} ({\bf 0}, C)}\times D_d \longrightarrow \overline{B_{\left\| \cdot \right\|} ({\bf 0}, C)}$ and $F _\rho^f: \overline{B_{\left\| \cdot \right\|} ({\bf 0}, C_f)}\times D_d \longrightarrow \overline{B_{\left\| \cdot \right\|} ({\bf 0}, C_f)}$. Then, both $F _\rho $ and $F _\rho^f $ have the ESP and have unique FMP associated filters ${U}_\rho:({D}_d)^{\mathbb{Z}_{-}}\longrightarrow K _C$ and ${U}_\rho^f:({D}_d)^{\mathbb{Z}_{-}}\longrightarrow K_{C _f}$, respectively. The state map $\mathcal{F} _\rho ^f  :f ^\ast  \left(\overline{B_{\left\| \cdot \right\|} ({\bf 0}, C_f)}\right)\times  D _d \longrightarrow f ^\ast  \left(\overline{B_{\left\| \cdot \right\|} ({\bf 0}, C_f)}\right)$ is isomorphic to the restricted version of $F _\rho^f $, also has the ESP and  a FMP associated filter  $\mathcal{U} _\rho ^f: ({D}_d)^{\mathbb{Z}_{-}}\longrightarrow \left(f ^\ast  \left(\overline{B_{\left\| \cdot \right\|} ({\bf 0}, C_f)}\right)\right)^{\mathbb{Z}_{-}}$.
The state map $\mathcal{F} _\rho ^f  $ and the filter $\mathcal{U} _\rho ^f $  are called the {\it JL projected} versions of $ {F} _\rho$ and  $ {U} _\rho $, respectively. 
\item [(iv)] In the hypotheses of the previous point, for any ${\bf z} \in (D_d)^{\mathbb{Z}_{-}} $ and $t \in \mathbb{Z}_{-} $:
\begin{equation}
\label{bound error reduced F 1}
\left\|U_{\rho}({\bf z})_t- \mathcal{U}_{\rho} ^f({\bf z})_t\right\|\leq \epsilon ^{1/2}CM_QC _Q\frac{(1+\vertiii{f}^2)^{1/2}}{1- \rho},
\end{equation}
where $M_Q$ and $C _Q $ are given by \eqref{definition of M} and \eqref{definition of C_Q}, respectively. Alternatively, it can also be shown that:
\begin{equation}
\label{bound error reduced F 2}
\left\|U_{\rho}({\bf z})_t- \mathcal{U}_{\rho} ^f({\bf z})_t\right\|\leq \epsilon \frac{CM_Q^2C _Q ^2}{1- \rho}.
\end{equation}
\item [(v)] Let $R> {\rm max}\{1/\vertiii{f}^2,1\}   $ and set $\rho=1/(R\vertiii{f} ^2) $. Then, the elements in the set $Q$ can be chosen so that the bounds in  \eqref{bound error reduced F 1} and \eqref{bound error reduced F 2} reduce to 
\begin{eqnarray}
 & &\!\!\!\!\!\!\!\!\!\!\!\!\!\!\!\epsilon ^{1/2}N^{3/4}C  \left(1+\vertiii{f}^2\right)^{1/2}\frac{R\vertiii{f}^2}{R\vertiii{f}^2-1} \quad \mbox{and}\label{reduced bounds}\\
 & &\epsilon NC \frac{R\vertiii{f}^2}{R\vertiii{f}^2-1},\label{reduced bounds 2}
\end{eqnarray}
respectively. 
\end{description}
\end{theorem}

\subsection{The Johnson-Lindenstrauss reduced  SigSAS system}

We now use the previous theorem to spell out the Johnson-Lindenstrauss projected version of  SigSAS approximations and to establish error bounds analogous to those introduced in \eqref{reduced bounds} and \eqref{reduced bounds 2}. Given that Theorem \ref{theorem Johnson-Lindenstrauss projection of state-space dynamics} is formulated using the one and the two-norms in Euclidean spaces and Proposition \ref{proposition The SigSAS system} defines the 
SigSAS system on a tensor space endowed with an unspecified cross-norm, we notice that those two frameworks can be matched by using the norms $\left\|\cdot \right\| $ and $\left\|\cdot \right\|_1 $ in $T^{l+1} (\mathbb{R}^{p+1})$ given by
\begin{equation*}
\left\|\mathbf{v}\right\| ^2:=\sum _{i _1, \ldots, i_{l+1}=1}^{p+1} \lambda _{i _1, \ldots, i_{l+1}} ^2,  \  \left\|\mathbf{v}\right\| ^2_1:=\sum _{i _1, \ldots, i_{l+1}=1}^{p+1} |\lambda _{i _1, \ldots, i_{l+1}}|,
\end{equation*}
with $\mathbf{v}=\sum _{i _1, \ldots, i_{l+1}=1}^{p+1} \lambda _{i _1, \ldots, i_{l+1}} \mathbf{e}_{i _1}\otimes\cdots \otimes\mathbf{e}_{i _{l+1}}$ and $\left\{\mathbf{e}_{i _1}\otimes\cdots \otimes\mathbf{e}_{i _{l+1}}\right\}_{i _1, \ldots, i_{l+1}\in \left\{1, \ldots, p+1\right\}}$ the canonical basis in $T^{l+1} (\mathbb{R}^{p+1})$. It is easy to check that these two norms are crossnorms and that $\left\|\cdot \right\| $ is the norm associated to the inner product defined by the extension by bilinearity of the assigment
\begin{equation*}
\langle\mathbf{e}_{i _1}\otimes\cdots \otimes\mathbf{e}_{i _{l+1}}, \mathbf{e}_{j _1}\otimes\cdots \otimes\mathbf{e}_{j _{l+1}}\rangle:= \delta_{i _1 j _1} \cdots\delta_{i _{l+1} j _{l+1}}, 
\end{equation*}
that makes $(T^{l+1} (\mathbb{R}^{p+1}), \langle\cdot , \cdot \rangle)$ into a Hilbert space, a feature that is needed to use the Johnson-Lindenstrauss Lemma.

\begin{corollary}
\label{JL reduced SigSAS}
Let $M>0$ and let $(F^{{\rm SigSAS}}_{\lambda,l , p}, W)$  be the  SigSAS system that approximates a causal and TI filter $U: K _M \longrightarrow \ell_{-}^{\infty}(\mathbb{R}^m) $,  as introduced in Theorem \ref{SigSAS approximation}. Let $N:= (p+1)^{l+1}$, $\widetilde{M } $ as in \eqref{contraction constant}, and let $0< \epsilon< 1 $. Let $f : \mathbb{R}^N \longrightarrow \mathbb{R}^k  $ be a JL map that satisfies \eqref{JL distances condition}, where the dimension $k$ has been chosen to make $f$ generically surjective. Then, for any $R> {\rm max}\{1/\vertiii{f}^2, 1/ (\widetilde{M}\vertiii{f}^2), 1\}   $,  $\lambda:= 1/ (R\widetilde{M}\vertiii{f}^2)$,  and $L$ as in \eqref{definition of L}, there exists a JL reduced version $\mathcal{F}^{{\rm SigSAS}}_{\lambda,l , p, f}: f ^\ast  \left(\overline{B_{\left\|\cdot \right\|}({\bf 0}, L_f)}\right) \times [-M,M] \longrightarrow  f ^\ast  \left(\overline{B_{\left\|\cdot \right\|}({\bf 0}, L_f)}\right) $ of ${F}^{{\rm SigSAS}}_{\lambda,l , p}:  \overline{B_{\left\|\cdot \right\|}({\bf 0}, L)} \times [-M,M] \longrightarrow  \overline{B_{\left\|\cdot \right\|}({\bf 0}, L)}$, with $L _f:= \widetilde{M}\vertiii{f}/ \left(1- \lambda \widetilde{M}\vertiii{f}^2\right) $,
that has the ESP and a unique FMP associated filter $\mathcal{U}^{{\rm SigSAS}}_{\lambda,l , p, f}: K _M \longrightarrow \left(f ^\ast  \left(\overline{B_{\left\|\cdot \right\|}({\bf 0}, L_f)}\right)\right)^{\Bbb Z _-}$. Moreover, we have that
\begin{multline}
\label{error bound jl projection sigsas1}
\left\| W{U}^{{\rm SigSAS}}_{\lambda,l , p}({\bf z})_t- {\cal W}\mathcal{U}^{{\rm SigSAS}}_{\lambda,l , p, f}({\bf z})_t\right\|\\
\leq \vertiii{W}\epsilon^{\frac{1}{2}}N^{\frac{3}{4}}(1+\vertiii{f}^2)^{\frac{1}{2}} \frac{\widetilde{M}R ^2\vertiii{f}^4}{(R\vertiii{f}^2-1)^2}, 
\end{multline}
\begin{multline}
\label{error bound jl projection sigsas2}
\left\| W{U}^{{\rm SigSAS}}_{\lambda,l , p}({\bf z})_t- {\cal W}\mathcal{U}^{{\rm SigSAS}}_{\lambda,l , p, f}({\bf z})_t\right\|\\
\leq \vertiii{W} \epsilon N \frac{\widetilde{M} R ^2\vertiii{f}^4}{(R\vertiii{f}^2-1)^2},
\end{multline}
for any ${\bf z} \in K _M $ and $t \in \mathbb{Z}_{-}$, and where ${\cal W}:=W \circ i _k \in \mathbb{M}_{m,k}$, with $i _k: f ^\ast \circ f (T^{l+1}(\mathbb{R}^{p+1}))\hookrightarrow T^{l+1}(\mathbb{R}^{p+1})$ the inclusion.
\end{corollary}

This  result shows that causal and time-invariant filters can be approximated by JL reduced SigSAS systems. The goal in the following paragraphs consists in showing that such systems are just {\it   SAS systems with randomly drawn matrix coefficients} and, additionally, in precisely spelling out the law of their entries. {These facts show precisely} that a large class of filters can be learnt just by randomly generating a SAS and by tuning   a linear readout layer for each individual filter that needs to be approximated. We emphasize that the JL reduced randomly generated SigSAS system is the same for the entire class of FMP filters that are being approximated and that only the linear readout depends on the individual filter that needs to be learnt, which amounts to the strong universality property that we discussed in the Introduction and in Section \ref{The signature state-affine system (SigSAS)}. As in Remark \ref{strong but}, we recall that the quality of the approximation using a JL reduced random SigSAS system may change from filter to filter because of the dependence on the sequence $w ^U $ in the bound \eqref{volterra representation truncated sigsas} and the presence of the linear readout $W$ in \eqref{error bound jl projection sigsas1} and \eqref{error bound jl projection sigsas2}.

The next statement needs the following fact that is known in the literature as {\it  Gordon's Theorem} (see \cite[Theorem 5.32]{vershynin:random:matrices} and references therein): given a random matrix $A \in \mathbb{M}_{n,m} $ with standard Gaussian IID entries, we have that 
\begin{equation}
\label{latala ineq}
{\rm E} \left[\vertiii{A}\right]\leq \sqrt{n}+\sqrt{m}.
\end{equation}

Additionally, the element  $\widehat{ {\bf z}} ^0 \in T^{l+1}(\mathbb{R}^{p+1}) $ introduced in \eqref{definition z0} for the construction of the SigSAS system will be chosen in a specific randomized way in this case. Indeed, this time around, we replace \eqref{definition z0} by 
\begin{equation}
\label{definition z0 randomized}
\widehat{{\bf z}} ^0=r\sum_{i \in I _0} z^{i-1}\mathbf{e}_{1}\otimes \cdots \otimes\mathbf{e}_{1}\otimes\mathbf{e}_{i},
\end{equation}
where $r$ is a Rademacher random variable that is chosen independent from all the other random variables that will appear in the different constructions. If we take in $ T^{l+1}(\mathbb{R}^{p+1}) $ the canonical basis in lexicographic order, the element $\widehat{{\bf z}} ^0 $ can be written as the image of a linear map as
\begin{equation}
\label{definition z0 randomized matrix form}
\widehat{{\bf z}} ^0= r C^{I _0}(1, z, \ldots, z ^p)^{\top}, \quad \mbox{with} \quad
\end{equation} 
\begin{equation*}
C^{I _0}:=
\left(
\begin{array}{c}
S^c\\
\mathbb{O}_{(p+1)\left((p+1) ^l-1\right), p+1}
\end{array}
\right) \in \mathbb{M}_{(p+1)^{l+1},p+1},
\end{equation*}
and $S^c \in \mathbb{M}_{p+1}$ a diagonal selection matrix with the elements given by $S^c_{ii}=1  $ if $i \in I _0 $, and $S^c _{ii}=0$ otherwise.

\begin{theorem}
\label{JL reduced system forgotten above}
Let $M>0 $,  let $\widetilde{M } $ as in \eqref{contraction constant},  $l,p, k \in \mathbb{N} $, and define $N:= (p+1)^{l+1}$, $N_0:= (p+1)^{l}$. Consider a SigSAS state map $F^{{\rm SigSAS}}_{\lambda,l , p}:T^{l+1}(\mathbb{R}^{p+1}) \times [-M,M] \longrightarrow T^{l+1}(\mathbb{R}^{p+1}) $ of the type introduced in \eqref{sigsas state function} and defined by choosing the non-homogeneous term $\widehat{ {\bf z}} ^0  $ as in \eqref{definition z0 randomized}.
Let now $f : \mathbb{R}^N \longrightarrow \mathbb{R}^k  $ be a JL projection randomly drawn  according to \eqref{choice for JL 1/k}. 
Let $\delta> 0  $ be small enough so that 
\begin{equation}
\label{choice of lambda0}
\lambda _0:= \frac{\delta }{2\widetilde{M}} \sqrt{\frac{k}{N _0}} <\min \left\{\dfrac{1}{\widetilde{M}}, \dfrac{1}{\widetilde{M}\vertiii{f}^2},1\right\}.
\end{equation}
Then, the JL reduced version $\mathcal{F}^{{\rm SigSAS}}_{\lambda _0,l , p, f}$ of ${F}^{{\rm SigSAS}}_{\lambda_0,l , p}$ has the ESP and the FPM with probability at least $1- \delta $ and, in the limit $N _0 \rightarrow  \infty $,
it is isomorphic to the family of randomly generated SAS systems ${F}^{{\rm SigSAS}}_{\lambda _0,l , p,f}$ with states in $\mathbb{R}^k $ and given by
\begin{equation}
\label{final reduced with random matrices}
F^{{\rm SigSAS}}_{\lambda_0,l , p, f}(\mathbf{x}, z):=\sum_{i=1}^{p+1}z ^{i-1}A _i \mathbf{x} +B \left(1,z ,\cdots,z ^p\right)^{\top},
\end{equation}
where $A _1, \ldots, A_{p+1} \in \mathbb{M} _k$ and $B \in \mathbb{M}_{k, p+1} $ are random matrices whose entries are drawn according to: 
\begin{equation}
\label{law random matrix sigsas}
(A _1)_{j,m}, \ldots, (A _{p+1})_{j,m}\sim {\rm N}\left(0, \frac{\delta ^2}{4k \widetilde{M}^2 }\right), 
\end{equation}
\begin{equation}
\label{law random matrix sigsas bs}
B_{j,m}\sim 
\left\{
\begin{array}{ll}
{\rm N} \left(0, \frac{1}{k}\right) &\  \mbox{if $m \in I _0 $,}\\
0 &\  \mbox{otherwise.}
\end{array}
\right.
\end{equation}
All the entries in the matrices $A _1, \ldots, A_{p+1} $ are independent random variables. The entries in the matrix $B$ are independent from each other and  they are decorrelated and asymptotically independent (in the limit as $N _0 \rightarrow  \infty $) from those in $A _1, \ldots, A_{p+1} $.
\end{theorem}

We conclude with a result that uses in a combined manner the SigSAS Approximation Theorem \ref{SigSAS approximation}  with its JL reduction in Corollary \ref{JL reduced SigSAS}, as well as its  SAS characterization with random coefficients in Theorem \ref{JL reduced system forgotten above}. This statement shows that in order to approximate a large class of sufficiently regular FMP filters with uniformly bounded inputs, it suffices to  randomly generate a common SAS system for all of them and  to tune a linear readout for each different filter in that class that needs to be approximated.

\begin{theorem}
\label{final approximation theorem}
Let $M, L>0 $ and let $U:K _M \subset  \ell_{-}^{\infty}(\mathbb{R})\longrightarrow K_L \subset \ell_{-}^{\infty}(\mathbb{R}^m) $ be a causal and time-invariant fading memory filter that satisfies the hypotheses in Theorem \ref{SigSAS approximation}. Fix now $l,p,k \in \mathbb{N} $ and $\delta> 0 $ small enough so that \eqref{choice of lambda0} holds. Construct now the SAS system with states in $\mathbb{R}^k $ given by
\begin{equation}
\label{final reduced with random matrices forgotten}
F^{{\rm SigSAS}}_{\lambda_0,l , p, f}(\mathbf{x}, z)=\sum_{i=1}^{p+1}z ^{i-1}A _i \mathbf{x} +B \left(1,z ,\cdots,z ^p\right)^{\top},
\end{equation}
with matrix coefficients randomly generated  according to the laws spelled out in \eqref{law random matrix sigsas} and \eqref{law random matrix sigsas bs}. 

If $p$ and $l $ are large enough, then the SAS system $F^{{\rm SigSAS}}_{\lambda_0,l , p, f} $ has the ESP and the FPM with probability at least $1- \delta $. In that case $F^{{\rm SigSAS}}_{\lambda_0,l , p, f} $ has a filter $U ^{{\rm SigSAS}}_{\lambda_0,l , p, f}$ associated and there exists a monotonically decreasing sequence $w ^U $ with zero limit and a linear map $\overline{W} \in L(\mathbb{R}^k, \mathbb{R} ^m)$ such that {for any ${\bf z}\in\widetilde{B}_M$} it holds that
\footnotesize
\begin{multline}
\label{final theorem volterra representation truncated sigsas}
\left\|U ({\bf z})_t- \overline{W}U ^{{\rm SigSAS}}_{\lambda_0,l , p, f}({\bf z})_t \right\|\\
\leq w ^U _l+L \left(1- \frac{\left\|{\bf z}\right\|_{\infty}}{M}\right)^{-1} \left(\frac{\left\|{\bf z}\right\|_{\infty}}{M}\right)^{p+1}+ I_{l,p},
\end{multline}
\normalsize
where $I_{l,p} $ is either
\begin{align}
\label{expression ilp}
I_{l,p}&:=\vertiii{W}\epsilon^{\frac{1}{2}}N^{\frac{3}{4}} \widetilde{M} \dfrac{\left(1+\vertiii{f}^2\right)^{\frac{1}{2}}}{\left(1-\dfrac{\delta}{2} \sqrt{\dfrac{k}{N_0}}\right)^2} \quad  \mbox{or} \nonumber \\ I_{l,p}&:=\vertiii{W}  \epsilon N \widetilde{M} \dfrac{1}{\left(1-\dfrac{\delta}{2} \sqrt{\dfrac{k}{N_0}}\right)^2}.
\end{align}
In these expressions ${W}\in L(T^{l+1}(\mathbb{R}^{p+1}), \mathbb{R}^m)$ is a linear map such that $\overline{W}={W} \circ f^\ast$, $N= (p+1)^{l+1} $, $\widetilde{M} $ is defined in \eqref{contraction constant}, and $0< \epsilon< 1$ satisfies \eqref{JL dimension condition} with $n$ replaced by $N $.

\end{theorem}

\section{Conclusion}

Reservoir computing capitalizes on the remarkable fact that there are learning systems that attain universal approximation properties without requiring that all their parameters are estimated using a supervised learning procedure. These untrained parameters are most of the time randomly generated and it is only an output layer that  needs to be estimated using a simple functional prescription. This phenomenon has been explained for static (extreme learning machines  \cite{Huang2006}) and dynamic (echo state networks \cite{hart:ESNs, RC12}) neural paradigms and its performance has been quantified using mostly probabilistic methods. 

In this paper, we have concentrated on a different class of reservoir computing systems, namely the state-affine (SAS) family. The SAS class was introduced and proved universal in \cite{RC6} and we have shown here that the possibility of randomly constructing these systems and at the same time preserving their approximation properties is of geometric nature. The rationale behind our description relies on the following points:
\begin{itemize}[leftmargin=*]
\item Any analytic filter can be represented as a Volterra series expansion. When this filter is additionally of fading memory type, the truncation error can be easily quantified.
\item Truncated Volterra series admit a natural state-space representation with linear observation equation in a conveniently chosen tensor space. The state equation of this representation has a strong universality property whose unique solution can be used to approximate any analytic fading memory filter just by modifying the linear observation equation. We refer to this strongly universal filter as the SigSAS system.
\item The random projections of the SigSAS system yield SAS systems with randomly generated coefficients in a potentially much smaller dimension which approximately preserve the good properties of the original SigSAS system. The loss in performance that one incurs  because of the projection mechanism can be quantified using the Johnson-Lindenstrauss Lemma.
\end{itemize}

These observations collectively show that {\it SAS reservoir systems with randomly chosen coefficients exhibit excellent empirical performances in the learning of fading memory input/output systems because they approximately correspond to very high-degree Volterra series expansions of those systems}. That degree is actually of the order of the exponential of the dimension of the randomly generated SAS system used.


%

\appendix

\subsection{Proof of Theorem \ref{volterra approximation}}
\footnotesize
The representation \eqref{volterra representation} is a straightforward multivariate generalization of Theorem 29 in \cite{RC9}.  {For any ${\bf z}\in \widetilde{B}_M $ and any $p,l\in \mathbb{N}$ define
\begin{equation*}
U^{l,p}({\bf z})_t:=\sum_{j=1}^{p}\sum_{m _1=- l}^0 \cdots\sum_{m _j=- l}^0 g _j(m _1, \ldots, m _j)({\bf z}_{ m _1+ t}\otimes \cdots \otimes{\bf z}_{ m _j+ t}).
\end{equation*}
Now, for any ${\bf z}\in \widetilde{B}_M $ and $t _1, t _2 \in \mathbb{Z}_{-}  $  such that  $t _2\leq t _1 $, define the sequence ${\bf z}^{t _1}_{t _2} \in  \widetilde{B}_M $ by ${\bf z}^{t _1}_{t _2}:= (\ldots, {\bf 0},{\bf z}_{t _2}, \ldots, {\bf z}_{t _1}) $. Additionally, for any ${\bf u} \in (\mathbb{R}^d)^{\mathbb{Z}_-}$ and any ${\bf z} \in (\mathbb{R}^d)^{\mathbb{N}^+}$, the symbol $  {\bf uz}^1 _t  \in (\mathbb{R}^d)^{\mathbb{Z}_-}$, $t \in \mathbb{N}^+$,  denotes the concatenation of the left-shifted vector $\mathbf{u} $ with the truncated vector ${\bf z}^1 _t:= ({\bf z}  _1, \ldots, {\bf z} _t)$ obtained out of ${\bf z} $. With this notation, we now show \eqref{volterra representation truncated}. By the triangle inequality and the time-invariance of $U$, for any ${\bf z}\in \widetilde{B}_M $ we have }
\begin{multline}
\label{intermediate for w}
\!\!\!\!\!\!\!\!\left\|U ({\bf z})_t- U^{l,p}({\bf z})_t\right\|\leq \left\|U ({\bf z})_t- U^{l,\infty}({\bf z})_t\right\|+\left\|U^{l,\infty} ({\bf z})_t- U^{l,p}({\bf z})_t\right\|\\
=\left\|\sum_{j=1}^{\infty}\sum_{m _1=- \infty}^{-l-1} \cdots\sum_{m _j=- \infty}^{-l-1} g _j(m _1, \ldots, m _j)({\bf z}_{ m _1+ t}\otimes \cdots \otimes{\bf z}_{ m _j+ t})\right\|\\
+\left\|U({\bf z}_{-l+t}^t) _0-U^{\infty,p}({\bf z}_{-l+t}^t) _0\right\|\\
=\left\|U({\bf z}_{- \infty}^{-l-1+t}{\bf 0}_{l+1})_0\right\|+\left\|U({\bf z}_{-l+t}^t) _0-U^{\infty,p}({\bf z}_{-l+t}^t) _0\right\|,
\end{multline}
where the symbol ${\bf 0}_{l+1}$ stands for a $l+1 $-tuple of the element ${\bf 0} \in {\Bbb R}^d $. The second summand of this expression can be bounded using the Taylor bound provided in  \cite[Theorem 29]{RC9}. As to the first summand, we shall use the input forgetting property that the filter $U$ exhibits since, by hypothesis,  has the FMP. More specifically, if we apply Theorem  6 in \cite{RC9} to the FMP filter $U:K _M \longrightarrow \ell_{-}^{\infty}(\mathbb{R}^m) $, we can conclude the existence of a montonically decreasing sequence $w ^U $ with zero limit such that 
\begin{equation*}
\left\|U({\bf z}_{- \infty}^{-l-1+t}{\bf 0}_{l+1})_0\right\|=\left\|U({\bf z}_{- \infty}^{-l-1+t}{\bf 0}_{l+1})_0-U({\bf 0})_0\right\|\leq w ^U_{l},
\end{equation*}
for any $l\in \mathbb{N}$.
These two arguments substituted in \eqref{intermediate for w} yield the bound in \eqref{volterra representation truncated}. \quad $\blacksquare$
\normalsize

\subsection{Proof of Proposition \ref{proposition The SigSAS system}} 
\footnotesize The map $F^{{\rm SigSAS}}_{\lambda,l , p}:T^{l+1}(\mathbb{R}^{p+1}) \times [-M, M] \longrightarrow T^{l+1}(\mathbb{R}^{p+1}) $ is clearly continuous and, additionally, it is a contraction in its first component. Indeed, let $\mathbf{x}_1, \mathbf{x} _2 \in T^{l+1}(\mathbb{R}^{p+1}) $  and let $z \in [-M,M] $ be arbitrary. Notice first that
\begin{equation}
\label{bound for z tilde}
\left\|\widetilde{{\bf z}}\right\|= \left\|\sum _{i=1}^{p+1} z ^{i-1}\mathbf{e} _i \right\|\leq 1+M+ \cdots + M ^p= \frac{1-M^{p+1}}{1-M}=: \widetilde{M}.
\end{equation}
Now, since we are using a crossnorm in $T^{l+1}(\mathbb{R}^{p+1})$, we have that,
\begin{multline*}
\left\|F^{{\rm SigSAS}}_{\lambda,l , p}(\mathbf{x}_1,z)-F^{{\rm SigSAS}}_{\lambda,l , p}(\mathbf{x}_2,z)\right\|= \lambda\left\|\pi _l(\mathbf{x} _1- \mathbf{x} _2)\otimes \widetilde{{\bf z}}\right\|\\
=\lambda\left\|\pi _l(\mathbf{x} _1- \mathbf{x} _2)\right\| \left\|\widetilde{{\bf z}}\right\|.
\end{multline*}
If we use in this equality the relation \eqref{bound for z tilde} and the fact that $\vertiii{\pi _l}=1$, we can conclude that:
\begin{equation}
\label{contractivity condition fsas}
\left\|F^{{\rm SigSAS}}_{\lambda,l , p}(\mathbf{x}_1,z)-F^{{\rm SigSAS}}_{\lambda,l , p}(\mathbf{x}_2,z)\right\|\leq \lambda \widetilde{M} \left\|\mathbf{x} _1- \mathbf{x} _2\right\|.
\end{equation}
The hypothesis $ \lambda < (1-M)/(1-M^{p+1})= 1/\widetilde{M} $ implies that $F^{{\rm SigSAS}}_{\lambda,l , p} $ is a contraction and establishes \eqref{contraction constant}. Additionally, $\left\|\widehat{ {\bf z}} ^0\right\|= \left\|\sum_{i \in I _0} z^{i-1}\mathbf{e}_{1}\otimes \cdots \otimes\mathbf{e}_{1}\otimes\mathbf{e}_{i}\right\|\leq 1+M+ \cdots + M ^p= \widetilde{M}$, which implies that 
\begin{equation}
\label{to prove that it restricts}
\left\|F^{{\rm SigSAS}}_{\lambda,l , p} ({\bf 0}, z)\right\|\leq \widetilde{M}, \quad \mbox{for all} \quad z \in [-M,M],
\end{equation}
and hence by \cite[Remark 2]{RC10} we can conclude that  $F^{{\rm SigSAS}}_{\lambda,l , p} $ restricts to a map $F^{{\rm SigSAS}}_{\lambda,l , p}: \overline{B _{\left\|\cdot \right\|}({\bf 0},L)} \times [-M, M] \longrightarrow \overline{B _{\left\|\cdot \right\|}({\bf 0},L)}  $, for any $L\geq \widetilde{M}/ (1- \lambda \widetilde{M})$. Finally, the contractivity condition established in \eqref{contractivity condition fsas} and \cite[Theorem 12]{RC9}, imply that the corresponding state system has the ESP and the FMP. We now show that its unique solution is given by \eqref{solution sigsas}. First, it is easy to see that by iterating the recursion \eqref{sigsas recursion} twice and three times,  one obtains:
\begin{multline*}
\mathbf{x} _t =  \lambda^2 \pi_l \left(\pi _l (\mathbf{x}_{t-2})\otimes \widetilde{{\bf z}} _{t-1}\right)\otimes \widetilde{{\bf z}} _{t}+ \lambda \pi_l(\widehat{ {\bf z}} ^0 _{t-1})\otimes \widetilde{{\bf z}} _t+ \widehat{{\bf z}} _t^0\\
	= \lambda^3 \pi_l(\pi_l \left(\pi _l (\mathbf{x}_{t-3})\otimes \widetilde{{\bf z}} _{t-2}\right)\otimes \widetilde{{\bf z}} _{t-1})\otimes \widetilde{{\bf z}} _{t}\\
+ \lambda^2 \pi _l(\pi_l(\widehat{ {\bf z}} ^0 _{t-2})\otimes \widetilde{{\bf z}} _{t-1})\otimes \widetilde{{\bf z}} _{t}+ \lambda \pi_l(\widehat{ {\bf z}} ^0 _{t-1})\otimes \widetilde{{\bf z}} _t+ \widehat{{\bf z}} _t^0.
\end{multline*}
More generally, after $l+1 $ iterations one obtains,
\begin{multline*}
\label{solution sigsas}
\mathbf{x} _t=  \lambda ^{l+1}
\underbrace{\pi _l( \pi_l( \cdots (\pi_l(}_\text{$l+1$-times}\mathbf{x}_{t-(l+1)})\otimes\widetilde{{\bf z}}_{t-l}) \otimes \cdots )\otimes \widetilde{{\bf z}}_{t-1}) \otimes\widetilde{{\bf z}}_{t}\\
+ \lambda ^l
\underbrace{\pi _l( \pi_l( \cdots (\pi_l(}_\text{$l$-times}\widehat{{\bf z}}^0_{t-l})\otimes\widetilde{{\bf z}}_{t-(l-1)}) \otimes \cdots )\otimes \widetilde{{\bf z}}_{t-1}) \otimes\widetilde{{\bf z}}_{t}\\\
+ \cdots+ \lambda \pi_l(\widehat{{\bf z}}^0_{t-1})\otimes \widetilde{ {\bf z}} _t+ \widehat{{\bf z}}^0_{t}.
\end{multline*}
Consequently, in order to establish \eqref{solution sigsas} it suffices to show that 
\begin{equation}
\label{point to be proves sigsas solution}
 \lambda ^{l+1}
\underbrace{\pi _l( \pi_l( \cdots (\pi_l(}_\text{$l+1$-times}\mathbf{x}_{t-(l+1)})\otimes\widetilde{{\bf z}}_{t-l}) \otimes \cdots )\otimes \widetilde{{\bf z}}_{t-1}) \otimes\widetilde{{\bf z}}_{t}=\frac{\lambda^{l+1}}{1- \lambda}\widehat{{\bf z}} _t.
\end{equation}
We show this equality by writing
\begin{equation}
\label{convention x with a}
\mathbf{x} _t=\sum_{i _1, \ldots, i_{l+1}=1}^{p+1}a_{i _1, \ldots, i_{l+1}}^t \mathbf{e}_{i _1 }\otimes\cdots\otimes\mathbf{e}_{i _{l+1} },
\end{equation}
for some coefficients $a_{i _1, \ldots, i_{l+1}}^t \in \mathbb{R}  $  that by \eqref{sigsas recursion} and the assumption that $1 \in I _0 $, satisfy that
\begin{equation*}
a ^t_{1, \ldots, 1}= \lambda a ^{t-1}_{1, \ldots, 1}+1, \quad \mbox{for any} \quad t \in \mathbb{Z}_{-}.
\end{equation*}
{This recursion  can be rewritten for any $r\in \mathbb{N}$ and $t \in \mathbb{Z}$ as
\begin{equation*}
a ^t_{1, \ldots, 1}=\sum_{j=0}^{r - 1}\lambda^j + \lambda ^r a ^{t-r}_{1, \ldots, 1}.
\end{equation*}
Since by hypothesis the parameter $\lambda<1 $ and, additionally, we just showed that $\|\mathbf{x}_t\| \le L$, for all $t \in \mathbb{Z}$, with $L\ge \widetilde{M}/ (1- \lambda \widetilde{M})$,   this equation has a unique solution given by}
\begin{equation}
\label{solution small recursion}
a ^t_{1, \ldots, 1}=\frac{1}{1- \lambda}, \quad \mbox{for all} \quad t \in \mathbb{Z}_{-}.
\end{equation}
Now, notice that using \eqref{convention x with a}, we can write:
\begin{multline*}
\pi _l(\mathbf{x}_{t-(l+1)})\otimes\widetilde{{\bf z}}_{t-l}\\
=\sum_{i _2, \ldots, i_{l+1}, j _1=1}^{p+1}a_{1,i _2, \ldots, i_{l+1}}^{t-(l+1)}z_{t-l}^{j _1-1} \mathbf{e}_{i _2 }\otimes\cdots\otimes\mathbf{e}_{i _{l+1} }\otimes\mathbf{e}_{j _1}.
\end{multline*}

If we repeat this procedure $l+1 $ times, we obtain that 
\begin{multline*}
\underbrace{\pi _l( \pi_l( \cdots (\pi_l(}_\text{$l+1$-times}\mathbf{x}_{t-(l+1)})\otimes\widetilde{{\bf z}}_{t-l}) \otimes \cdots )\otimes \widetilde{{\bf z}}_{t-1}) \otimes\widetilde{{\bf z}}_{t}\\
=\sum_{j _1, \ldots, j _{l+1}=1}^{p+1}a_{1, \ldots,1}^{t-(l+1)}z_{t-l}^{j _1-1}z_{t-(l-1)}^{j _2-1}\cdots z_{t}^{j _{l+1}-1} \mathbf{e}_{j _1}\otimes\cdots\otimes\mathbf{e}_{ j _{l+1}}\\
= a_{1, \ldots,1}^{t-(l+1)}\widehat{{\bf z}} _t
=\frac{1}{1- \lambda}\widehat{{\bf z}} _t,
\end{multline*}
where the last equality is a consequence of \eqref{solution small recursion}. This identity proves \eqref{point to be proves sigsas solution}. $\blacksquare$

\subsection{Proof of Theorem \ref{SigSAS approximation}} 
{\footnotesize It is a straightforward corollary of Theorem \ref{volterra approximation} and of the expression \eqref{solution sigsas} of the filter $U^{{\rm SigSAS}}_{\lambda,l , p}$. The linear map $W$ is constructed by matching the coefficients $g _j(m _1, \ldots, m _j) $ of the truncated Volterra series representation of $U$ up to polynomial degree $p$ with the terms of the filter $U^{{\rm SigSAS}}_{\lambda,l , p}({\bf z})_t$ in the canonical basis of $T^{l+1}(\mathbb{R}^{p+1}) $. More specifically, $W \in L(T^{l+1}(\mathbb{R}^{p+1}), \mathbb{R} ^m)$ is the linear map that satisfies:
\begin{equation}
\label{determines W}
WU^{{\rm SigSAS}}_{\lambda,l , p}({\bf z})_t= \sum_{j=1}^{p}\sum_{m _1=- l}^0 \cdots\sum_{m _j=- l}^0 g _j(m _1, \ldots, m _j){z}_{ m _1+ t} \cdots {z}_{ m _j+ t},
\end{equation}
for any ${\bf z} \in K _M $, $t \in \mathbb{Z}_{-} $, where the right hand side of this equality is the truncated Volterra series expansion of $U$, available by Theorem \ref{volterra approximation}. The equality \eqref{determines W} does determine $W $ because by \eqref{transformation for monomials}, it is equivalent to: 
\begin{multline*}
\sum_{i _1, \ldots, i _{l+1}=1}^{p+1} WA_{\lambda,l , p} \left(\mathbf{e}_{i _1}\otimes \cdots \otimes\mathbf{e}_{i _{l+1}}\right) z_{t-l}^{i_1-1} \cdots z_{t}^{i_{l+1}-1}\\
=\sum_{j=1}^{p}\sum_{m _1=- l}^0 \cdots\sum_{m _j=- l}^0 g _j(m _1, \ldots, m _j){z}_{ m _1+ t} \cdots {z}_{ m _j+ t}.
\end{multline*}
Since this equality between polynomials has to hold for any ${\bf z} \in K _M $ and $t \in \mathbb{Z}_{-} $, we can conclude that the matrix coefficients on both sides have to coincide. This implies that, in particular, for any $i _1, \ldots, i _{l+1} \in \left\{1, \ldots, p+1\right\}$,
\begin{equation}
\label{components W}
WA_{\lambda,l , p} \left(\mathbf{e}_{i _1}\otimes \cdots \otimes\mathbf{e}_{i _{l+1}}\right)= \sum_{I_{i _1, \ldots, i _{l+1}}}g _j(m _1, \ldots, m _j) \in \mathbb{R}^m,
\end{equation}
where $
I_{i _1, \ldots, i _{l+1}}= \left\{(j , m _1, \ldots, m _j)\right\} $ is the set of indices with $j \in \left\{1, \ldots, p\right\}$, $ m _i \in \left\{-l, \ldots, 0\right\}$,  and ${z}_{ m _1} \cdots {z}_{ m _j}=z_{-l}^{i_1-1} \cdots z_{0}^{i_{l+1}-1}$.
As \eqref{components W} specifies the image of a basis by the map $WA_{\lambda,l , p} $ and $A_{\lambda,l , p}$  is invertible, then
\eqref{components W} and consequently \eqref{determines W} fully determine $W $. The bound in \eqref{volterra representation truncated sigsas} is then a consequence of \eqref{determines W} and \eqref{volterra representation truncated} in Theorem \ref{volterra approximation}. \quad $\blacksquare$}
\normalsize

\subsection{Proof of Lemma \ref{definition of Q norm}}
\footnotesize
\noindent\textbf{(i)} It is obvious that if $\mathbf{v}={\bf 0} $ then $\left\|\mathbf{v} \right\| _Q=0 $ and that $\left\|\lambda\mathbf{v} \right\| _Q=| \lambda| \left\|\mathbf{v} \right\| _Q$, for all $\lambda \in \mathbb{R}  $  and $\mathbf{v} \in {\rm span}\left\{Q\right\} $. Let now $\mathbf{w} _1, \mathbf{w} _2 \in {\rm span}\left\{Q\right\} $ and $C_q:={\rm Card} \, Q$. Given that,
\begin{multline*}
\!\!\!\!\!\!{\rm inf} \left\{\sum_{j=1}^{C_q }| \lambda _j^1+ \lambda _j^2|\mid \sum_{j=1}^{C_q } \lambda _j^1 \mathbf{v} _j= \mathbf{w}_1, \sum_{j=1}^{C_q } \lambda _j^2 \mathbf{v} _j= \mathbf{w}_2, \mathbf{v} _j \in Q\right\}\\
\geq {\rm inf} \left\{\sum_{j=1}^{C_q }| \lambda _j|\mid \sum_{j=1}^{C_q } \lambda _j \mathbf{v} _j=  \mathbf{w}_1+\mathbf{w}_2, \mathbf{v} _j \in Q\right\},
\end{multline*}
we can conclude that
\begin{multline*}
\left\|\mathbf{w} _1+ \mathbf{w} _2\right\| _Q\leq\\
{\rm inf} \left\{\sum_{j=1}^{C_q }| \lambda _j^1+ \lambda _j^2|\mid \sum_{j=1}^{C_q } \lambda _j^1 \mathbf{v} _j= \mathbf{w}_1, \sum_{j=1}^{C_q } \lambda _j^2 \mathbf{v} _j= \mathbf{w}_2, \mathbf{v} _j \in Q\right\} \\
\leq{\rm inf} \left\{\sum_{j=1}^{C_q }| \lambda _j^1|+ |\lambda _j^2|\mid \sum_{j=1}^{C_q } \lambda _j^1 \mathbf{v} _j= \mathbf{w}_1, \sum_{j=1}^{C_q } \lambda _j^2 \mathbf{v} _j= \mathbf{w}_2, \mathbf{v} _j \in Q\right\}\\
= 
\left\|\mathbf{w} _1\right\| _Q+ \left\|\mathbf{w} _2\right\| _Q,
\end{multline*}
which establishes the {triangle} inequality and hence shows that $\left\|\cdot \right\| _Q $  is a seminorm. Suppose now that $M_Q < \infty $  and let $\mathbf{v} \in {\rm span}\left\{Q\right\}  $ such that $\left\|\mathbf{v} \right\| _Q=0 $. By the approximation property of the infimum, for any $\epsilon >0 $ there exist $\lambda_1, \ldots, \lambda_{C_q} \in \mathbb{R}$ such that $\sum_{j=1}^{C_q } \lambda _j \mathbf{v} _j= \mathbf{v} $ and  $0\leq \sum_{j=1}^{C_q }| \lambda _j|< \epsilon $. This inequality implies that
\begin{equation}
\label{recycle ineq}
\left\|\mathbf{v}\right\|= \left\|\sum_{j=1}^{C_q } \lambda _j \mathbf{v} _j\right\|\leq M_Q \sum_{j=1}^{C_q }| \lambda _j|<M_Q \epsilon.
\end{equation}
Since $M_Q$ is finite and $\epsilon>0 $ can be made arbitrarily small, this inequality implies that $\left\|\mathbf{v}\right\|=0 $  and hence $\mathbf{v}= {\bf 0} $, necessarily, which proves that $\left\|\cdot \right\| _Q $ is  a norm in this case.

Since the first inequality \eqref{recycle ineq} holds for any $\mathbf{v} \in {\rm span}\left\{Q\right\}  $, the statement in part {\bf (ii)} follows (when $M _Q  $ is not finite we use the convention that $\infty \cdot 0 = \infty $).  Part {\bf (iii)} is obvious. \quad $\blacksquare$
\normalsize

\subsection{Proof of Proposition \ref{quasi projection proposition}} 
\footnotesize
Since $V$  and $\mathbb{R} ^k $ are Hilbert spaces, the parallelogram law holds for the associated norms and hence, for any $\mathbf{v} _1, \mathbf{v} _2 \in Q $,
\begin{multline}
\label{paralelo and fstarf}
\langle\mathbf{v} _1, \mathbf{v} _2- f ^\ast \circ f (\mathbf{v} _2)\rangle= \langle\mathbf{v}_1, \mathbf{v} _2\rangle- \langle f(\mathbf{v} _1), f(\mathbf{v} _2) \rangle\\
= \frac{1}{4}\left( \left\|\mathbf{v} _1+ \mathbf{v} _2\right\|^2- \left\|\mathbf{v} _1- \mathbf{v} _2\right\|^2\right)\\
- \frac{1}{4}\left( \left\|f(\mathbf{v} _1)
+f( \mathbf{v} _2)\right\|^2- \left\|f(\mathbf{v} _1)- f(\mathbf{v} _2)\right\|^2\right)\\
=\frac{1}{4}\left( \left\|\mathbf{v} _1-(- \mathbf{v} _2)\right\|^2- \left\|\mathbf{v} _1- \mathbf{v} _2\right\|^2\right)\\
- \frac{1}{4}\left( \left\|f(\mathbf{v} _1)-f(- \mathbf{v} _2)\right\|^2- \left\|f(\mathbf{v} _1)- f(\mathbf{v} _2)\right\|^2\right)\\
\leq \frac{\epsilon}{4} \left(\left\|\mathbf{v} _1- \mathbf{v} _2\right\|^2+ \left\|\mathbf{v} _1+ \mathbf{v} _2\right\|^2\right)= \frac{\epsilon}{2} \left(\left\|\mathbf{v} _1\right\|^2+ \left\|\mathbf{v} _2\right\|^2\right),
\end{multline}
where in the inequality in the last line we used the JL property \eqref{JL distances condition} together with the hypothesis $-Q=Q $. Let now $\mathbf{w}_1=\sum_{i=1}^{{\rm Card}\, Q}\lambda ^1 _i \mathbf{v} _i, \mathbf{w} _2=\sum_{i=1}^{{\rm Card}\, Q}\lambda ^2 _i \mathbf{v} _i \in {\rm span}\left\{Q\right\}  $. Then, by \eqref{paralelo and fstarf}:
\begin{multline*}
\left|\langle\mathbf{w} _1, \mathbf{w} _2- f ^\ast \circ f (\mathbf{w} _2)\rangle\right|= \left|
\sum_{i,j=1}^{{\rm Card}\, Q}\lambda ^1 _i\lambda ^2 _j \langle\mathbf{v} _i, \mathbf{v} _j- f ^\ast \circ f (\mathbf{v} _j)\rangle\right|\\
\leq 
 \sum_{i,j=1}^{{\rm Card}\, Q}|\lambda ^1 _i||\lambda ^2 _j| \frac{\epsilon}{2} \left(\left\|\mathbf{v} _i\right\|^2+ \left\|\mathbf{v} _j\right\|^2\right)
\leq \sum_{i=1}^{{\rm Card}\, Q}|\lambda ^1 _i|  \sum_{j=1}^{{\rm Card}\, Q} |\lambda ^2 _j| \epsilon M_Q ^2.
\end{multline*}
Since this inequality holds true for any linear decomposition of $\mathbf{w} _1 , \mathbf{w}_2\in {\rm span}\left\{Q\right\} $, we can take infima on its right hand side with respect to those decompositions, which clearly implies \eqref{epsilon condition ffstar}. \quad $\blacksquare$
\normalsize

\subsection{Proof of Theorem \ref{theorem Johnson-Lindenstrauss projection of state-space dynamics}} 

\footnotesize
\noindent {\bf (i)} We  show that when condition \eqref{esp condition in f} holds, then $F _\rho^f  $ is a contraction on the first entry. Let $\mathbf{x} _1, \mathbf{x} _2 \in \mathbb{R} ^k  $ and let ${\bf z} \in D _d$, then
\begin{multline*}
\left\|F _\rho^f(\mathbf{x} _1, {\bf z})-F _\rho^f(\mathbf{x} _2, {\bf z})\right\|\\
= \left\|f \left(F _\rho(f ^\ast (\mathbf{x}_1), {\bf z})\right)-f \left(F _\rho(f ^\ast (\mathbf{x}_2), {\bf z})\right)\right\|
\leq \rho\vertiii{f}\vertiii{f^\ast }\left\|\mathbf{x} _1- \mathbf{x} _2\right\|.
\end{multline*}
The claim follows from this inequality, the equality $\vertiii{f}=\vertiii{f^\ast }$,  and condition \eqref{esp condition in f}.

\medskip

\noindent {\bf (ii)} The proof is straightforward. The only point that needs to be emphasized is that $(f ^\ast ) ^{-1}: V _k \longrightarrow \mathbb{R}^k $ is well-defined because since $f$ is surjective, then $f ^\ast : \mathbb{R}^k\longrightarrow V _k $ is necessarily injective.

\medskip

\noindent {\bf (iii)}  First of all, the existence of the restricted versions of $F _\rho $ and $F _\rho ^f  $ to compact state-spaces and the fact that these maps are contractions on the first entry with contraction rates $\rho $ and $\rho \vertiii{f}^2 $, respectively, implies by  \cite[Theorem 7, part~(i)]{RC9} that they have the ESP and  associated FMP filters $U _\rho $ and $U _\rho ^f $. The statement about the JL-projected state map $\mathcal{F} ^f _\rho $  and its associated filter $\mathcal{U} _\rho ^f  $ is a straightforward consequence of the fact that the restricted linear map $f ^\ast : \mathbb{R} ^k \longrightarrow V_k $ is a state-map equivariant linear isomorphism between $F ^f _\rho $  and $\mathcal{F} ^f _\rho $ and of the properties of this kind of maps (see, for instance, \cite[Proposition 2.3]{RC15}). 

\medskip

\noindent {\bf (iv)} Let ${\bf z} \in (D_d)^{\mathbb{Z}_{-}} $ and $t \in \mathbb{Z}_{-} $ arbitrary. Then, using \eqref{definition frho}, we have
\begin{multline}
\label{to be bounded temp}
\left\|U_{\rho}({\bf z})_t- \mathcal{U}_{\rho} ^f({\bf z})_t\right\|= \left\|F _\rho(U_{\rho}({\bf z})_{t-1}, {\bf z} _t)-\mathcal{F} _\rho(\mathcal{U}_{\rho} ^f({\bf z})_{t-1}, {\bf z} _t)\right\|\\=
\|F _\rho(U_{\rho}({\bf z})_{t-1}, {\bf z} _t)-F _\rho(\mathcal{U}_{\rho} ^f({\bf z})_{t-1}, {\bf z} _t)\\
+F _\rho(\mathcal{U}_{\rho} ^f({\bf z})_{t-1}, {\bf z} _t)  -\mathcal{F} _\rho(\mathcal{U}_{\rho} ^f({\bf z})_{t-1}, {\bf z} _t)\|\\ \leq
\rho \left\|U_{\rho}({\bf z})_{t-1}-\mathcal{U}_{\rho} ^f({\bf z})_{t-1}\right\|+ \left\|(\mathbb{I}_N- f ^\ast \circ f)(F _\rho(\mathcal{U}_{\rho} ^f({\bf z})_{t-1}, {\bf z} _t)) \right\|.
\end{multline}
The bounds in \eqref{bound error reduced F 1} and \eqref{bound error reduced F 2} are obtained by bounding the last expression in  \eqref{to be bounded temp} in two different fashions. First, if we use \eqref{epsilon condition ffstar} and the hypothesis on $Q$ being a spanning set of $\mathbb{R}^N$, we have that:
\begin{multline}
\label{to replace by other one}
\rho \left\|U_{\rho}({\bf z})_{t-1}-\mathcal{U}_{\rho} ^f({\bf z})_{t-1}\right\|+ \left\|(\mathbb{I}_N- f ^\ast \circ f)F _\rho(\mathcal{U}_{\rho} ^f({\bf z})_{t-1}, {\bf z} _t) \right\|\\\leq
\rho \left\|U_{\rho}({\bf z})_{t-1}-\mathcal{U}_{\rho} ^f({\bf z})_{t-1}\right\|\\
+ \epsilon ^{1/2}M_Q \left\|(\mathbb{I}_N- f ^\ast \circ f)(F _\rho(\mathcal{U}_{\rho} ^f({\bf z})_{t-1}, {\bf z} _t)) \right\|_Q^{1/2} \left\|F _\rho(\mathcal{U}_{\rho} ^f({\bf z})_{t-1}, {\bf z} _t)\right\|_Q^{1/2}\\\leq
\rho \left\|U_{\rho}({\bf z})_{t-1}-\mathcal{U}_{\rho} ^f({\bf z})_{t-1}\right\|+ \epsilon^{1/2}CM_QC _Q \left(1+ \vertiii{f} ^2\right)^{1/2}.
\end{multline}
If we now iterate the  procedure in \eqref{to be bounded temp} on the first summand of this expression, we obtain
\begin{multline}
\label{iterative bounding procedure}
\left\|U_{\rho}({\bf z})_t- \mathcal{U}_{\rho} ^f({\bf z})_t\right\|= \left\|F _\rho(U_{\rho}({\bf z})_{t-1}, {\bf z} _t)-\mathcal{F} _\rho(\mathcal{U}_{\rho} ^f({\bf z})_{t-1}, {\bf z} _t)\right\|\\\leq
\rho \left(\rho\left\|U_{\rho}({\bf z})_{t-2}-\mathcal{U}_{\rho} ^f({\bf z})_{t-2}\right\|+ \epsilon^{1/2}CM_QC _Q \left(1+ \vertiii{f} ^2\right)^{1/2}\right)\\
+ \epsilon^{1/2}CM_QC _Q \left(1+ \vertiii{f} ^2\right)^{1/2}\\=
\rho^2 \left\|U_{\rho}({\bf z})_{t-2}-\mathcal{U}_{\rho} ^f({\bf z})_{t-2}\right\|+ (1+ \rho)\epsilon^{1/2}CM_QC _Q \left(1+ \vertiii{f} ^2\right)^{1/2}\\\leq
\rho^j \left\|U_{\rho}({\bf z})_{t-j}-\mathcal{U}_{\rho} ^f({\bf z})_{t-j}\right\|\\
+ (1+ \rho+ \rho ^2+ \cdots \rho^{j-1})\epsilon^{1/2}CM_QC _Q \left(1+ \vertiii{f} ^2\right)^{1/2}.
\end{multline}
As by hypothesis $\rho< 1 $, we can take the limit $j \rightarrow \infty $ in this expression, which yields \eqref{bound error reduced F 1}. In order to obtain \eqref{bound error reduced F 2} it suffices to replace the use of \eqref{epsilon condition ffstar} in \eqref{to replace by other one} by that of \eqref{inequ with norm cq}.

\medskip

\noindent {\bf (v)} First of all, note that for any $R> {\rm max}\{1/\vertiii{f}^2,1\}   $, the contraction parameter $\rho=1/(R\vertiii{f}^2) $ satisfies the condition \eqref{esp condition in f}. Set now $Q:= \left\{\pm \mathbf{e} _1, \ldots, \pm \mathbf{e}_N \right\} $. It is easy to see that with this choice, the norm $\left\|\cdot \right\| _Q $ introduced in Lemma \ref{definition of Q norm} satisfies that $\left\|\cdot \right\|_Q= \left\|\cdot \right\|_1$ and that $M _Q=1 $. If we now recall that $\left\|\cdot \right\|\leq \left\|\cdot \right\|_1\leq \sqrt{N} \left\|\cdot \right\| $ and that $(1/\sqrt{N})\vertiii{\cdot }\leq \vertiii{\cdot }_1\leq \sqrt{N} \vertiii{\cdot } $, we can rewrite the inequality \eqref{to replace by other one} as
\begin{multline*}
\left\|U_{\rho}({\bf z})_t- \mathcal{U}_{\rho} ^f({\bf z})_t\right\|\leq
\rho \left\|U_{\rho}({\bf z})_{t-1}-\mathcal{U}_{\rho} ^f({\bf z})_{t-1}\right\|\\
+ \epsilon ^{1/2}\vertiii{\mathbb{I}_N- f ^\ast \circ f}_1^{1/2} \left\|F _\rho(\mathcal{U}_{\rho} ^f({\bf z})_{t-1}, {\bf z} _t)\right\|_1\\\leq
\rho \left\|U_{\rho}({\bf z})_{t-1}-\mathcal{U}_{\rho} ^f({\bf z})_{t-1}\right\|+ \epsilon ^{1/2}N^{3/4}C\vertiii{\mathbb{I}_N- f ^\ast \circ f}^{1/2}\\\leq
\epsilon ^{1/2}N^{3/4}C\vertiii{\mathbb{I}_N- f ^\ast \circ f}^{1/2}\frac{1}{1- \rho}\\
\leq 
\epsilon ^{1/2}N^{3/4}C(1+\vertiii{f}^2)^{1/2}\frac{R \vertiii{f}^2}{R \vertiii{f}^2-1},
\end{multline*}
where in the passage from the second to the third line we just used the same iterative bounding procedure as in \eqref{iterative bounding procedure}. The bound in \eqref{reduced bounds 2} is obtained by using 
the inequality in \eqref{variation for later} adapted to our particular choice of $Q$, according to which, for any $\mathbf{v} \in \mathbb{R}^N $, we have that
\begin{multline*}
\left\|(\mathbb{I}_N- f ^\ast \circ f )(\mathbf{v})\right\|^2\leq \epsilon \left\|(\mathbb{I}_N- f ^\ast \circ f )(\mathbf{v})\right\| _1 \left\|\mathbf{v}\right\|_1\\
\leq \epsilon\sqrt{N} \left\|(\mathbb{I}_N- f ^\ast \circ f) (\mathbf{v})\right\|\left\|\mathbf{v}\right\|_1,
\end{multline*}
which implies that 
\begin{equation}
\label{iffwith1}
\left\|(\mathbb{I}_N- f ^\ast \circ f) (\mathbf{v})\right\|\leq \epsilon\sqrt{N} \left\|\mathbf{v}\right\|_1.
\end{equation}
Consequently, by \eqref{to be bounded temp} and \eqref{iffwith1}, we have
\begin{multline*}
\left\|U_{\rho}({\bf z})_t- \mathcal{U}_{\rho} ^f({\bf z})_t\right\| \leq
\rho \left\|U_{\rho}({\bf z})_{t-1}-\mathcal{U}_{\rho} ^f({\bf z})_{t-1}\right\|\\
+ \left\|(\mathbb{I}_N- f ^\ast \circ f)(F _\rho(\mathcal{U}_{\rho} ^f({\bf z})_{t-1}, {\bf z} _t)) \right\|\\\leq 
\rho \left\|U_{\rho}({\bf z})_{t-1}-\mathcal{U}_{\rho} ^f({\bf z})_{t-1}\right\|+ \epsilon\sqrt{N}\left\|F _\rho(\mathcal{U}_{\rho} ^f({\bf z})_{t-1}, {\bf z} _t) \right\|_1\\
\leq \epsilon NC \frac{1}{1- \rho}=\epsilon NC \frac{R\vertiii{f}^2}{R\vertiii{f}^2-1}. \quad \blacksquare
\end{multline*}
\normalsize

\subsection{Proof of Corollary \ref{JL reduced SigSAS}} 

\footnotesize
It consists of just applying the JL projection Theorem \ref{theorem Johnson-Lindenstrauss projection of state-space dynamics} to the SigSAS system introduced in Proposition \ref{proposition The SigSAS system}. First of all, by \eqref{contraction constant}, the state map $F^{{\rm SigSAS}}_{\lambda,l , p}:  \overline{B_{\left\|\cdot \right\|}({\bf 0}, L)} \times [-M,M] \longrightarrow  \overline{B_{\left\|\cdot \right\|}({\bf 0}, L)} $ is continuous and a contraction on the state variable, with contraction constant $\rho:= \lambda \widetilde{M} $. Now, if we fix $R> {\rm max}\{1/\vertiii{f}^2, 1/ \widetilde{M}\vertiii{f}^2, 1\}   $, then $\lambda= 1/ R\widetilde{M}\vertiii{f}^2 $ satisfies all the necessary constraints in the statement of Proposition \ref{proposition The SigSAS system}, as well as the condition $\rho<1/\vertiii{f}^2$, as required in \eqref{esp condition in f}.

Therefore, given a JL map $f$, part {\bf (i)} of Theorem \ref{theorem Johnson-Lindenstrauss projection of state-space dynamics}  shows the existence of a contractive state map $F^{{\rm SigSAS}}_{\lambda,l , p, f} $ with constant $\lambda \widetilde{M}\vertiii{f}^2 $ defined by $F^{{\rm SigSAS}}_{\lambda,l , p, f}(\mathbf{x}, z):= f \left(F^{{\rm SigSAS}}_{\lambda,l , p}( f ^\ast (\mathbf{x}),z)\right) $, $\mathbf{x} \in \mathbb{R}^k$, $z \in [-M,M] $. Following the same strategy as in \eqref{to prove that it restricts}, it is easy to see that $\left\|F^{{\rm SigSAS}}_{\lambda,l , p, f} ({\bf 0}, z)\right\|\leq \widetilde{M}\vertiii{f} $, for all  $z \in [-M,M]$, and hence by \cite[Remark 2]{RC10} we can conclude that  $F^{{\rm SigSAS}}_{\lambda,l , p, f} $ restricts to a map $F^{{\rm SigSAS}}_{\lambda,l , p,f }: \overline{B _{\left\|\cdot \right\|}({\bf 0},L _f)} \times [-M, M] \longrightarrow \overline{B _{\left\|\cdot \right\|}({\bf 0},L_f)}  $, with  $L_f= \widetilde{M}\vertiii{f}/ (1- \lambda \widetilde{M}\vertiii{f}^2)$. In those circumstances, part {\bf (iii)} of Theorem \ref{theorem Johnson-Lindenstrauss projection of state-space dynamics} allow us to conclude that the JL reduced SigSAS state map $\mathcal{F}^{{\rm SigSAS}}_{\lambda,l , p, f}: f ^\ast  \left(\overline{B_{\left\|\cdot \right\|}({\bf 0}, L_f)}\right) \times [-M,M] \longrightarrow  f ^\ast  \left(\overline{B_{\left\|\cdot \right\|}({\bf 0}, L_f)}\right) $ has the ESP and a unique FMP associated filter $\mathcal{U}^{{\rm SigSAS}}_{\lambda,l , p, f}: K _M \longrightarrow \left(f ^\ast  \left(\overline{B_{\left\|\cdot \right\|}({\bf 0}, L_f)}\right)\right)^{\Bbb Z _-}$. 

The error bounds in \eqref{error bound jl projection sigsas1}-\eqref{error bound jl projection sigsas2} are obtained out of \eqref{reduced bounds}-\eqref{reduced bounds 2}. Indeed, we first notice that by \eqref{definition of L} $\left\|F^{{\rm SigSAS}}_{\lambda,l , p}(\mathbf{x}, z)\right\| \le L $, with $L:= \widetilde{M}/ (1- \lambda \widetilde{M})$, for any  $\mathbf{x}\in \overline{B_{\left\|\cdot \right\|}({\bf 0}, L)} $ and $z \in [-M,M]$, and we can hence take $L$ as the constant $C$ in the bounds  \eqref{reduced bounds}-\eqref{reduced bounds 2}. Consequently, using the bound \eqref{reduced bounds}, we have
\begin{multline*}
\left\| W{U}^{{\rm SigSAS}}_{\lambda,l , p}({\bf z})_t- {\cal W}\mathcal{U}^{{\rm SigSAS}}_{\lambda,l , p, f}({\bf z})_t\right\|\\
\leq \vertiii{W}\left\| {U}^{{\rm SigSAS}}_{\lambda,l , p}({\bf z})_t-  \mathcal{U}^{{\rm SigSAS}}_{\lambda,l , p, f}({\bf z})_t\right\|\\
\leq \vertiii{W}\epsilon^{\frac{1}{2}}N^{\frac{3}{4}}(1+\vertiii{f}^2)^{\frac{1}{2}} \frac{R \vertiii{f}^2\widetilde{M}}{(R\vertiii{f}^2-1)(1- \lambda\widetilde{M})}\\
=\vertiii{W}\epsilon^{\frac{1}{2}}N^{\frac{3}{4}}(1+\vertiii{f}^2)^{\frac{1}{2}} \frac{\widetilde{M}R ^2\vertiii{f}^4}{(R\vertiii{f}^2-1)^2}.
\end{multline*}
The second inequality follows from  \eqref{reduced bounds 2}. \quad $\blacksquare$
\normalsize

\subsection{Proof of Theorem \ref{JL reduced system forgotten above}} 

\footnotesize
According to the third part of Theorem \ref{theorem Johnson-Lindenstrauss projection of state-space dynamics}, the JL reduction $\mathcal{F}^{{\rm SigSAS}}_{\lambda,l , p, f}$ of  $F^{{\rm SigSAS}}_{\lambda,l , p}(\mathbf{x}, z)=\lambda \pi_l (\mathbf{x})\otimes \widetilde{{\bf z}} + \widehat{{\bf z}} ^0 $ is isomorphic to the map $F^{{\rm SigSAS}}_{\lambda,l , p, f}$  given by 
\begin{equation}
\label{reduced no basis}
F^{{\rm SigSAS}}_{\lambda,l , p, f}(\mathbf{x}, z):=\lambda f(\pi_l (f ^\ast (\mathbf{x}))\otimes \widetilde{{\bf z}}) + f(\widehat{{\bf z}} ^0).
\end{equation}
Take now the canonical bases in $T^{l+1}(\mathbb{R}^{p+1})$ (using the lexicographic order) and $\mathbb{R}^k$ and let $S \in \mathbb{M}_{k,N}$, $N= (p+1)^{l+1}$, be  the random matrix associated to the JL projection $f$ in those bases. Let also $N_0:=(p+1)^l$, we separate $S$ in blocks and write:
\begin{equation}
\label{specs for S}
S= \left(S _1| \cdots |S_{p+1}\right), \enspace \mbox{with} \enspace S _1, \ldots, S_{p+1}\in \mathbb{M}_{k, N_0}, \,  S _{ij}\sim {\rm N}(0, 1/k).
\end{equation}
The order-lowering map $\pi_l:T^{l+1}(\mathbb{R}^{p+1}) \longrightarrow T^{l}(\mathbb{R}^{p+1})$ takes in these bases the matrix form 
\begin{equation}
\label{expression of capital pil}
\Pi _l= \left(\mathbb{I}_{N_0}| \mathbb{O}_{N_0,p N_0}\right)\quad \mbox{and} \quad \Pi _l S ^{\top}=S _1^{\top}.
\end{equation}
Consequently \eqref{reduced no basis} can be rewritten in matrix form as
\begin{equation}
\label{first matrix contact}
F^{{\rm SigSAS}}_{\lambda,l , p, f}(\mathbf{x}, z)= \lambda \sum_{i=1}^{p+1} z ^{i-1}S(\Pi _l(S ^{\top}( \mathbf{x})) \otimes \mathbf{e} _i)+ S \widehat{{\bf z}} ^0.
\end{equation}
We analyze this map by first spelling out the matrix associated to the linear assignment $\mathbf{x} \longmapsto \Pi _l(S ^{\top}( \mathbf{x})) \otimes \mathbf{e} _i $, $i =1, \ldots , p+1  $. It is easy to check that:
\begin{multline}
\label{matrix form main}
 \Pi _l(S ^{\top} (\cdot ) )\otimes \mathbf{e} _i=\\
\ 
\\
\left(
\begin{array}{l}
\left.
\begin{array}{cccc}
\mathbb{O}_{i-1,1}&\mathbb{O}_{i-1,1} &\cdots &\mathbb{O}_{i-1,1}\\
S_{1,1} & S_{2,1} &\cdots &S_{k,1 }\\
\mathbb{O}_{p-i+1,1}&\mathbb{O}_{p-i+1,1} &\cdots &\mathbb{O}_{p-i+1,1}
\end{array}
\right\} \ \mbox{Block  $1$}\\
\begin{array}{ccccccccccccc}
 &\vdots & \vdots & \vdots & \vdots &\vdots & \vdots & \vdots & \vdots & \vdots & \vdots & \vdots &\vdots \\
 &\vdots & \vdots & \vdots & \vdots &\vdots & \vdots & \vdots & \vdots & \vdots & \vdots & \vdots &\vdots
\end{array}\\
\left.
\begin{array}{cccc}
\mathbb{O}_{i-1,1}&\mathbb{O}_{i-1,1} &\cdots &\mathbb{O}_{i-1,1}\\
S_{1,(p+1)^l} & S_{2,(p+1)^l} &\cdots &S_{k,(p+1)^l }\\
\mathbb{O}_{p-i+1,1}&\mathbb{O}_{p-i+1,1} &\cdots &\mathbb{O}_{p-i+1,1}\
\end{array}
\right\} \ \mbox{Block  $(p+1)^l$}
\end{array}
\right).
\end{multline}
Using this matrix expression, it is easy to see that, for any $j,m \in \left\{1, \ldots ,k\right\} $:
\begin{multline}
\label{full multi terms}
\left(S( \Pi _l(S ^{\top} (\cdot ) )\otimes \mathbf{e} _i)\right)_{j,m}=\sum_{r=1}^{N}S _{j,r}( \Pi _l(S ^{\top} (\cdot )) \otimes \mathbf{e} _i)_{r,m }\\
=\sum _{n=1}^{N_0}S_{j, i+(n-1)(1+p)}S_{m,n}, \enspace {\rm with} \enspace N_0=(p+1)^l,
\end{multline}
and hence by \eqref{specs for S}, each of these entries is the sum of the products of two independent zero mean normal random variables with variance $1/k $, unless those two factors are identical, which can only happen whenever  $j=m $  and $i+(n-1)(1+p)=n $ simultaneously, which only holds for $i= 1$, $n=1 $, and   for the diagonal terms $j=m $. This implies that
\begin{equation}
\label{sum CLT 1}
\left(S( \Pi _l(S ^{\top} (\cdot ) )\otimes \mathbf{e} _1)\right)_{j,m}\sim
\left\{
\begin{array}{ll}
\sum_{n=1}^{N _0}a_{j,m,n}^1, &\mbox{when $j\neq m$,}\\
\\
\sum_{n=1}^{N _0-1}a_{j,m,n}^1+ b_{j,m}, &\mbox{otherwise,}
\end{array}
\right.
\end{equation}
where $b_{j,m}=(1/k) P_{j,m} $,  $a_{j,m,n}^1=(1/2k)(Q_{j,m,n}-R_{j,m,n}) $, and $P_{j,m}, Q_{j,m,n}, R_{j,m,n}\sim \chi ^2(1) $ are independent random variables. Analogously, for $i \in \left\{2, \ldots, p+1\right\}$,
\begin{equation}
\label{sum CLT 2}
\left(S( \Pi _l(S ^{\top} (\cdot ) )\otimes \mathbf{e} _i)\right)_{j,m}\sim
\sum_{n=1}^{N _0}a_{j,m,n}^i,
\end{equation}
with $a_{j,m,n}^i$ as above. We now study the matrix form of the summand $S \widehat{{\bf z}} ^0 $ in \eqref{first matrix contact}. First of all, by \eqref{definition z0 randomized matrix form},
\begin{equation*}
S \widehat{{\bf z}} ^0=r SC^{I _0}(1, z, \ldots, z ^p)^{\top},
\end{equation*}
and hence it can be written as $B \left(1,z ,\cdots,z ^p\right)^{\top}= S \widehat{{\bf z}} ^0 $, with $B \in \mathbb{M}_{k, p+1}$ the matrix with components
\begin{equation}
\label{distribution elements in B}
B_{j,m}=r \sum _{k=1}^N S_{jk}C^{I _0}_{km}=rS_{jm} {\bf 1}_{\left\{m \in I _0\right\}},
\end{equation}
which  \eqref{law random matrix sigsas bs}, as the product of a Gaussian with a Rademacher random variable is Gaussian distributed.

We now prove the claim in \eqref{law random matrix sigsas} regarding the matrices $A _1, \ldots, A_{p+1}$. First of all, \eqref{sum CLT 1} and \eqref{sum CLT 2} show that for each $i \in \left\{1, \ldots, p+1\right\} $, the entries $\left(S( \Pi _l(S ^{\top} (\cdot ) )\otimes \mathbf{e} _i)\right)_{j,m} $ of $\widetilde{A} _i= \lambda _0 S( \Pi _l(S ^{\top} (\cdot ) )\otimes \mathbf{e} _i)=  (\delta / 2\widetilde{M}) \sqrt{k/N _0}  S( \Pi _l(S ^{\top} (\cdot ) )\otimes \mathbf{e} _i)$ may be of two types
\begin{equation}
\label{sum CLT 3}
\left(\widetilde{A} _i\right)_{j,m}\sim
\left\{
\begin{array}{ll}
\frac{\delta\sqrt{k}}{2 \widetilde{M}}\sqrt{N _0}\left(\frac{\sum_{n=1}^{N _0-1}a_{j,m,n}^i}{N _0} + \frac{b_{j,m}}{N _0}\right),&\mbox{if $i=1, j=m $,}\\
\frac{\delta\sqrt{k}}{2 \widetilde{M}}\sqrt{N _0}\left(\frac{\sum_{n=1}^{N _0}a_{j,m,n}^i}{N _0} \right), &\mbox{otherwise.}
\end{array}
\right.
\end{equation}
As long as we are in the cases $j\neq m $ or simultaneously $i=1 $ and $j=m  $, these entries are the sum of IID mean zero random variables of variance $1/k ^2 $ and hence by the Lindeberg Central Limit Theorem, they converge in distribution  to mean zero Gaussian random variables   $(A _i)_{j,m}\sim {\rm N}\left(0, \frac{\delta ^2}{4k\widetilde{M}^2 }\right) $, as required. 

This straightforward argument cannot be used when $j=m $ and $i\geq 2 $ as, in that situation, the random variable $S_{m,i} $ appears in two different summands in the expression \eqref{full multi terms}. The claim in that case is proved by using a martingale central limit theorem. Indeed, consider the right hand side of the equality \eqref{full multi terms} for fixed $i \geq 2$ and $j=m$, that is,
\begin{align}
\label{eq:sum}
\sum_{n=1}^{N_0} S_{j,i+(n-1)(1+p)} S_{j,n}.
\end{align}
As the entries $S_{j,n}$ are independent zero mean normal random variables with variance $1/k$, we have,
\begin{equation}
\label{variance pathological}
\text{Var}\left(\sum_{n=1}^{N_0} S_{j,i+(n-1)(1+p)} S_{j,n}\right)=\frac{N_0}{k^2}.
\end{equation}
Notice first that for $s \geq l$, we have that
\[
{\rm E}\left[\frac{k}{\sqrt{N_0}}\sum_{n=1}^{s} S_{j,i+(n-1)(1+p)} S_{j,n} \Big| \mathcal{F}_l\right ]=\frac{k}{\sqrt{N_0}}\sum_{n=1}^{l} S_{j,i+(n-1)(1+p)} S_{j,n} ,
\]
where $\mathcal{F}_l$ is the $\sigma$-algebra generated by 
$\{ S_{j,i+(n-1)(1+p)}, S_{j,n} \,| \, n\leq l\}$. This equality is a consequence of the fact that for $s \geq l$ at most one of the terms in the product
\[
S_{j,i+(s-1)(1+p)} S_{j,s} 
\]
is $\mathcal{F}_l$-measurable and the other one is independent with zero expectation.
We shall apply Theorem 3.2 in \cite{hall:heyde:book} using as martingale differences the random variables $Y_{N_0,n} $ defined by
\[
Y_{N_0,n}:=\frac{k}{\sqrt{N_0}} S_{j,i+(n-1)(1+p)} S_{j,n}.
\]
It is clear that 
$
\underset{n}{\max} \{|Y_{N_0,n}| \} \to 0  
$
in probability as $N_0$ tends to $\infty$. Moreover,  ${\rm E}\left[\underset{n}{\max} \{Y_{N_0,n}^2\}\right]$ is bounded in $N_0$ and
\[
\underset {N_0\to \infty }{\operatorname {plim} }\sum_{n=1}^{N_0} Y_{N_0,n}^2 = 1.
\]
 Indeed,
\begin{align}
{\rm E}\left[\left( \sum_{n=1}^{N_0} Y_{N_0,n}^2- 1\right) ^2\right] &= 2\sum_{s<n}^{N_0} {\rm E}\left[\left( Y_{N_0,s}^2- \frac{1}{N_0}\right)\left( Y_{N_0,n}^2-  \frac{1}{N_0} \right) \right] \nonumber\\
&+ \sum_{n=1}^{N_0} {\rm E}\left[\left(Y_{N_0,n}^2- \frac{1}{N_0} \right)  ^2\right],\label{eqSum}
\end{align}
where for the second sum, using $\sum_{n=1}^{N_0}{\rm E}\left[ Y_{N_0,n}^4 \right] = N_0 \dfrac{k^4}{N_0^2}\dfrac{9}{k^4}={9}/{N_0}$ and that $Y_{N_0,n}$, $n=1,\ldots, N_0$, are random variables with  zero mean and variance  $1/N_0$, it holds that
\begin{align*}
\sum_{n=1}^{N_0} {\rm E}\left[\left(Y_{N_0,n}^2- \frac{1}{N_0} \right)  ^2\right] =8/N_0 \underset{N_0\rightarrow \infty}{\longrightarrow} 0.
\end{align*}
Notice now that in the first sum in \eqref{eqSum} the terms involving independent martingale differences vanish.  Hence, it is only the summands which correspond to the case when $Y_{N_0,s}$ and $Y_{N_0,n}$ are made of non-independent random variables that need to be treated separately. We notice that this happens precisely when $i +(s-1)(1+p) = n$ and write
\begin{align*}
2&\sum_{s<n}^{N_0} {\rm E}\Big[\left( Y_{N_0,s}^2- \frac{1}{N_0}\right)\left( Y_{N_0,n}^2-  \frac{1}{N_0} \right) \Big]\\
&=2\sum_{s<n}^{N_0} {\rm E}\Big[ Y_{N_0,s}^2 Y_{N_0,n}^2-  \frac{1}{N_0^2}  \Big]\\
&= 2 \sum_{s>1+ \frac{1-i}{p}}^{N_0} {\rm E}\Big[ \dfrac{k^4}{N_0^2}S_{j,i+(s-1)(1+p)}^4 S_{j,s}^2 S_{j,i+(i-1)(1+p)+(s-1)(1+p)^2}^2\\ &-  \frac{1}{N_0^2} \Big]\le2 (N_0 - 1)\left( \dfrac{k^4}{N_0^2} \dfrac{3}{k^2}\dfrac{1}{k^2} -  \frac{1}{N_0^2}\right) = 4 (N_0 - 1)/N_0^2 \underset{N_0\rightarrow \infty}{\longrightarrow} 0.
\end{align*}

Hence by 
Theorem 3.2 in \cite{hall:heyde:book}
$
Y_{N_0,n}
$
converges in distribution to a standard Gaussian random variable and the claim follows.

The mutual independence between the components of the different
matrices $A_i$ and within each individual matrix is obtained by using a
multivariate version of the martingale central limit theorem: the
arguments are fully in line with the previous application of the martingale
central limit theorem. The same applies for the statement about $ B$.

%

We conclude by showing that with the choice of $\lambda_0  $ in \eqref{choice of lambda0}, the reduced SAS system $F^{{\rm SigSAS}}_{\lambda_0,l , p, f}$ in  \eqref{final reduced with random matrices} has the ESP and the FMP with probability at least $1- \delta $. Let $p (z):= \sum_{i=1}^{p+1} z^{i-1}A _i $ and recall that by \cite[Proposition 14]{RC9}, the system $F^{{\rm SigSAS}}_{\lambda_0,l , p, f}$ has the ESP and the FMP whenever 
\begin{equation*}
M _p=\sup_{z \in [-M,M]} \left\{\vertiii{p(z)}\right\}<1.
\end{equation*}
Given that 
\begin{equation*}
M _p=\sup_{z \in [-M,M]} \left\{\vertiii{\sum_{i=1}^{p+1} z^{i-1}A _i}\right\}\leq 
\sum_{i=1}^{p+1} M^{i-1}\vertiii{A _i},
\end{equation*}
it is clear that the following inclusion of events holds $\left\{M _p\geq 1\right\} \subset \left\{\sum_{i=1}^{p+1} M^{i-1}\vertiii{A _i}\geq 1\right\}$. This implies that
\begin{multline*}
\mathbb{P}( M _p\geq 1 )\leq \mathbb{P}\left( \sum_{i=1}^{p+1} M^{i-1}\vertiii{A _i}\geq 1 \right)\\
\leq \sum_{i=1}^{p+1} M^{i-1} {\rm E}\left[\vertiii{A _i}\right]\leq \widetilde{M}2\sqrt{k} \frac{\delta}{2\sqrt{k}\widetilde{M}}= \delta,
\end{multline*}
as required. Note that in the second inequality we used Markov's inequality and that the third one is a consequence of Gordon's Theorem \eqref{latala ineq} and of \eqref{law random matrix sigsas}. \quad $\blacksquare$
\normalsize

\subsection{Proof of Theorem \ref{final approximation theorem}}

\footnotesize
First of all, given $l, p \in \mathbb{N} $, consider the SigSAS filter $U^{{\rm SigSAS}}_{\lambda _0,l , p} $ associated to the parameter $\lambda _0$ defined in \eqref{choice of lambda0} and the linear map $\overline{W} \in L(T^{l+1}(\mathbb{R}^{p+1}), \mathbb{R} ^m)$ whose existence for the approximation of $U$ is guaranteed by Theorem \ref{SigSAS approximation}. Consider now a JL projection $f: \mathbb{R}^{N } \longrightarrow \mathbb{R}^k $ that satisfies \eqref{JL distances condition}  with a parameter $0< \epsilon< 1$ that satisfies  \eqref{JL dimension condition} with $n$ replaced by $N $ and the value $k\in \mathbb{N} $ in the statement. Let now $F^{{\rm SigSAS}}_{\lambda_0,l , p, f} $ be the JL reduced SigSAS system  given by Corollary \ref{JL reduced SigSAS} and that asymptotically takes the form \eqref{final reduced with random matrices forgotten} because of \eqref{final reduced with random matrices} in Theorem \ref{JL reduced system forgotten above}. Let $W \in L(T^{l+1}(\mathbb{R}^{p+1}, \mathbb{R}^m)$ be the linear map in \eqref{volterra representation truncated sigsas} in  Theorem \ref{SigSAS approximation}. Define $\mathcal{W}:=W \circ i_k$ as in Corollary~\ref{JL reduced SigSAS} and $\overline{W}:=\mathcal{W} \circ f^\ast = W \circ f^\ast$. We now use the triangle inequality and the third part of Theorem~\ref{theorem Johnson-Lindenstrauss projection of state-space dynamics} to write:
\begin{multline*}
\left\|U ({\bf z})_t- \overline{W} U ^{{\rm SigSAS}}_{\lambda_0,l , p, f}({\bf z})_t \right\|\\
\leq 
\left\|U ({\bf z})_t- {W}U^{{\rm SigSAS}}_{\lambda _0,l , p}({\bf z})_t \right\|+ 
\left\|{W}U^{{\rm SigSAS}}_{\lambda _0,l , p} ({\bf z})_t- \overline{W} U ^{{\rm SigSAS}}_{\lambda_0,l , p, f}({\bf z})_t  \right\|\\\\
\leq 
\left\|U ({\bf z})_t- {W}U^{{\rm SigSAS}}_{\lambda _0,l , p}({\bf z})_t \right\|+ 
\left\|{W}U^{{\rm SigSAS}}_{\lambda _0,l , p} ({\bf z})_t- \mathcal{W}\mathcal{U}^{{\rm SigSAS}}_{\lambda_0,l , p, f}({\bf z})_t \right\|\\
\leq w ^U _l+L \left(1- \frac{\left\|{\bf z}\right\|_{\infty}}{M}\right)^{-1} \left(\frac{\left\|{\bf z}\right\|_{\infty}}{M}\right)^{p+1}+ I_{l,p}.
\end{multline*}
In the last inequality we have used Theorem \ref{SigSAS approximation} to bound the first summand and the last bounds stated in Corollary \ref{JL reduced SigSAS}  for the second one. More specifically, regarding the last point, we first note that the parameter $R $ in \eqref{error bound jl projection sigsas1} and \eqref{error bound jl projection sigsas2} is determined in this case by the definition of $\lambda _0 $ in  \eqref{choice of lambda0} and the equality 
\begin{equation*}
\frac{\delta }{2\widetilde{M}} \sqrt{\frac{k}{N _0}} =\frac{1}{R\widetilde{M}\vertiii{f}^2}
\end{equation*}
with $N_0 = (p+1)^l$. Hence, the bounds in \eqref{error bound jl projection sigsas1} and \eqref{error bound jl projection sigsas2} can be written as 
\begin{multline*}
\vertiii{W}\epsilon^{\frac{1}{2}}N^{\frac{3}{4}}(1+\vertiii{f}^2)^{\frac{1}{2}} \frac{\widetilde{M}R ^2\vertiii{f}^4}{(R\vertiii{f}^2-1)^2}\\
= \vertiii{W}\epsilon^{\frac{1}{2}}N^{\frac{3}{4}} \widetilde{M} \dfrac{\left(1+\vertiii{f}^2\right)^{\frac{1}{2}}}{\left(1-\dfrac{\delta}{2} \sqrt{\dfrac{k}{N_0}}\right)^2}\quad \mbox{and}
\end{multline*}
\begin{equation*}
\vertiii{W}\epsilon N \frac{\widetilde{M} R ^2\vertiii{f}^4}{(R\vertiii{f}^2-1)^2}=\vertiii{W}  \epsilon N \widetilde{M} \dfrac{1}{\left(1-\dfrac{\delta}{2} \sqrt{\dfrac{k}{N_0}}\right)^2}, 
\end{equation*}
respectively, which prove \eqref{expression ilp}. \quad $\blacksquare$

\normalsize


\section*{Acknowledgment}

CC acknowledges partial financial support from the Vienna Science and Technology Fund (WWTF) grant MA16-021, FWF  START Grant Y 1235. LGonon and JPO acknowledge partial financial support  coming from the Research Commission of the Universit\"at Sankt Gallen and the Swiss National Science Foundation (grant number 200021\_175801/1).  JPO acknowledges partial financial support of the French ANR ``BIPHOPROC" project (ANR-14-OHRI-0002-02). JT acknowledges support from the ETH Foundation and the Swiss National Science Foundation (grant number 179114, ``Machine Learning in Finance''). The authors thank the hospitality and the generosity of the FIM at ETH Zurich and the Division of Mathematical Sciences of the Nanyang Technological University, Singapore, where a significant portion of the results in this paper were obtained.

\ifCLASSOPTIONcaptionsoff
  \newpage
\fi



\bibliographystyle{IEEEtran}

\end{document}